\documentclass{article}

\PassOptionsToPackage{round}{natbib}

\usepackage[main, final]{neurips_2026}

\usepackage[utf8]{inputenc} 
\usepackage[T1]{fontenc}    
\usepackage{hyperref}       
\usepackage{url}            
\usepackage{booktabs}       
\usepackage{amsfonts}       
\usepackage{nicefrac}       
\usepackage{microtype}      
\usepackage{xcolor}         

\usepackage{amsmath}
\usepackage{amsthm}
\usepackage{amssymb}
\usepackage{enumitem}
\usepackage{array}
\usepackage{longtable}
\usepackage{tabularx}
\usepackage{float}
\usepackage{multirow}
\usepackage{tikz}
\usepackage{standalone}
\usetikzlibrary{positioning, arrows.meta, shapes.geometric, fit, backgrounds, calc}

\usepackage{algorithm}
\usepackage{algpseudocode}

\hypersetup{
    colorlinks=true,
    linkcolor=blue!70!black,
    citecolor=blue!70!black,
    urlcolor=blue!70!black
}

\newcommand{\Y}{\mathbf{Y}}
\title{Mamba Flow Matching Neural Processes: Linear-Time Inference for Irregularly Observed Spatial Fields}

\usepackage{xcolor}

\definecolor{revwine}{RGB}{150,0,40}

\author{%
  Cosmo Santoni\thanks{Corresponding author: \texttt{cosmo.santoni@imperial.ac.uk}} \quad Giovanni Charles \quad Timothy James Hitge \\
  \textbf{Oliver John Watson \quad Elizaveta Semenova} \\[4pt]
  Imperial College London
}

\begin{document}

\maketitle
\begin{abstract}
Interpolating a spatial field, such as daily maximum temperature, from sparse and irregularly located observations is a recurring task across the environmental and health sciences. The standard tool, Gaussian-process kriging, refits hyperparameters and solves an $O(N^3)$ system for every new field. When many related fields are observed, such as daily maps from one sensor network, neural processes (NPs) instead learn the interpolation map from past fields and predict a new field in a single forward pass. Joint-predictive variants, which sample whole fields rather than independent values at each location, typically rely on self-attention over context and query points, incurring $O(K^2)$ compute in the combined point count $K$ and capping practical use at a few thousand points. State-space models offer a linear-time alternative, but a direct state-space adaptation compresses the context into a single hidden state before reading queries, reducing the information available to the decoder. We propose MambaFlowNP, a joint-predictive NP with linear compute, trained with a conditional flow-matching objective. A multi-directional Mamba-2 scan reads context and query tokens in a single shared sequence, so each query's velocity is read from its own scan position rather than from a compressed context summary, and because queries only read the scan state, predictions are consistent across query sets. MambaFlowNP is competitive with the strongest attention-based joint-predictive baselines on synthetic Mat\'ern Gaussian-process benchmarks, and across regional and continental-scale NOAA temperature interpolation tasks it attains the lowest RMSE, the lowest CRPS and near-nominal coverage on every tier, while running $114$--$784{\times}$ faster than the strongest joint-predictive baseline. A single forward pass scales to $10^6$ points on one NVIDIA L40S GPU in under a second. This makes joint prediction computationally feasible for continental-scale, irregularly sampled sensor networks, with predictive quality evaluated up to $K{=}5{,}580$.
\end{abstract}

\section{Introduction}
\label{sec:intro}

Modelling continuous spatial fields from discrete, scattered sensor
data remains a fundamental challenge in spatial statistics and
probabilistic machine learning. Examples include inferring temperature
from weather stations~\citep{vaughan2022convolutional},
disease prevalence from sentinel sites~\citep{mappin2025mapping},
and rainfall between rain gauges~\citep{li2023rainfall}. Gaussian-process kriging is the standard approach, but it refits
hyperparameters and solves an $O(N^3)$ system for every new field. When many related fields are observed, as with daily readings from a
fixed sensor network, neural processes~\citep[NPs;][]{garnelo2018neural} cast this as amortised
inference, in which the network is trained on many datasets drawn from a data-generating process and, given a new
dataset, performs inference in a single forward pass without further
optimisation. An episode is one dataset together with a random split into context
observations and query locations at which values are to be
predicted. Factorised NPs infer the
parameters of an independent Gaussian distribution at each
query~\citep{garnelo2018conditional}, whilst latent-variable NPs
introduce a global latent variable~\citep{garnelo2018neural} through which all cross-query dependence
passes. In contrast, a fully
joint-predictive NP directly models the conditional distribution of
the values at all query locations together, via
autoregressive~\citep{nguyen2022transformer},
diffusion~\citep{dutordoir2023neural} or
flow~\citep{abuhamad2025flownp} parametrisations. Most existing
fully joint-predictive instances employ self-attention over the
concatenation of $N$ context and $M$ query locations, with compute
scaling as $O(K^2)$, where $K=N+M$. Flow Matching NP
(FNP)~\citep{abuhamad2025flownp} and Transformer NP
(TNPA)~\citep{nguyen2022transformer} are representative, and FNP exhausts GPU memory at $16{,}384$ points and TNPA at $65{,}536$
(Table~\ref{tab:scaling}). As an alternative to self-attention
transformers, state-space models (SSMs) such as
Mamba-2~\citep{dao2024mamba2} run a recurrent scan in $O(K)$ time,
FLOPs, and activation memory. However, a naive adaptation of these
models compresses the context into a single hidden state, a
bottleneck that we show degrades accuracy on synthetic and NOAA tasks
(\S\ref{sec:ablation-results}).

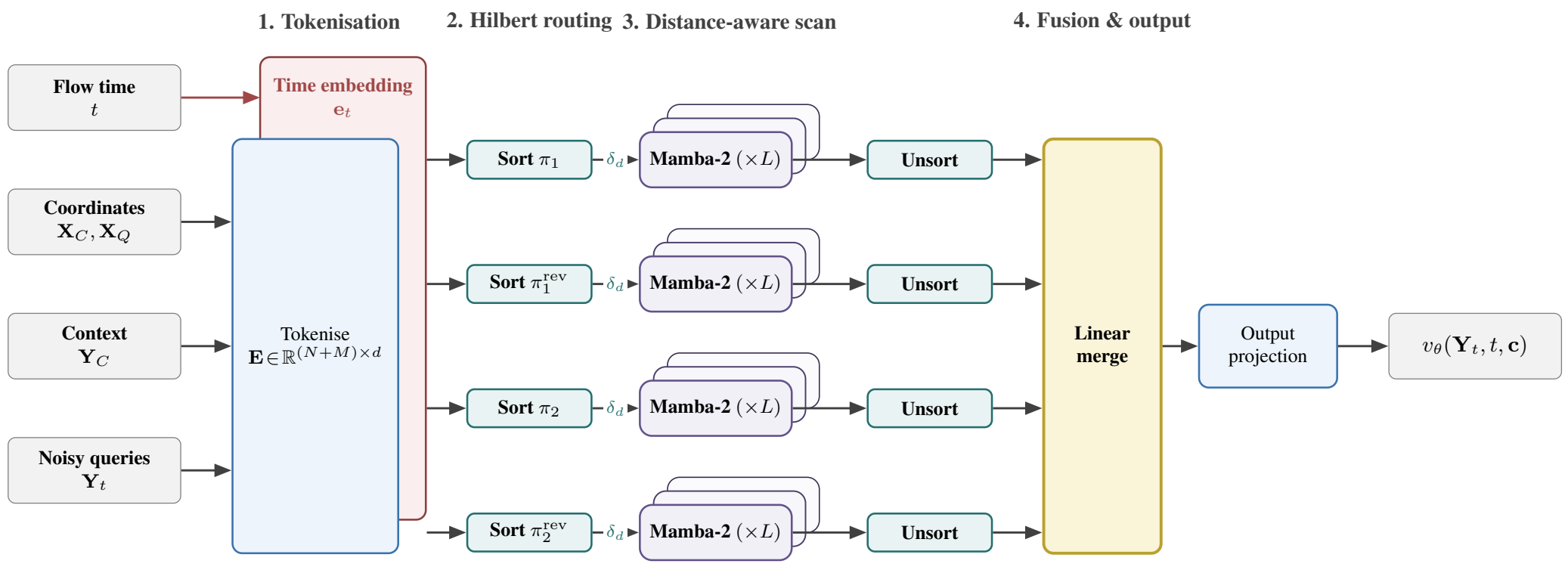
\begin{figure}[!htbp]
\centering
\resizebox{\linewidth}{!}{\usetikzlibrary{arrows.meta, calc, positioning}

\definecolor{myblue}{HTML}{4A90D9}
\definecolor{myteal}{HTML}{2E8B8B}
\definecolor{mypurple}{HTML}{7B68AE}
\definecolor{myyellow}{HTML}{E8C83A}
\definecolor{myred}{HTML}{C85A5A}

\begin{tikzpicture}[
    >={Stealth[scale=1.0, inset=0pt]},
    font=\sffamily\footnotesize,
    input node/.style={
        rectangle, rounded corners=3pt, draw=black!50, fill=gray!10,
        align=center, inner sep=4pt, font=\footnotesize\bfseries, text=black,
        minimum width=2.5cm, minimum height=0.95cm
    },
    process block/.style={
        rectangle, rounded corners=4pt, draw=myblue!80!black, fill=myblue!10, line width=0.8pt,
        align=center, inner sep=5pt, font=\footnotesize, text=black
    },
    sort block/.style={
        rectangle, rounded corners=3pt, draw=myteal!80!black, fill=myteal!10, line width=0.8pt,
        align=center, inner sep=3pt, font=\footnotesize\bfseries, text=black,
        minimum width=1.8cm, minimum height=0.55cm
    },
    mamba front/.style={
        rectangle, rounded corners=5pt, draw=mypurple!80!black, fill=mypurple!10, line width=0.8pt,
        align=center, inner sep=4pt, font=\footnotesize\bfseries, text=black,
        minimum width=2.2cm, minimum height=0.8cm
    },
    mamba mid/.style={
        rectangle, rounded corners=5pt, draw=mypurple!60!black, fill=mypurple!6, line width=0.6pt,
        minimum width=2.2cm, minimum height=0.8cm
    },
    mamba back/.style={
        rectangle, rounded corners=5pt, draw=mypurple!45!black, fill=mypurple!3, line width=0.5pt,
        minimum width=2.2cm, minimum height=0.8cm
    },
    fuse block/.style={
        rectangle, rounded corners=4pt, draw=myyellow!80!black, fill=myyellow!20, line width=1.2pt,
        align=center, inner sep=5pt, font=\footnotesize\bfseries, text=black
    },
    et back/.style={
        rectangle, rounded corners=4pt, draw=myred!70!black, fill=myred!10, line width=0.8pt,
        align=center, inner sep=5pt, font=\footnotesize\bfseries, text=black,
    },
    conn/.style={->, line width=0.9pt, color=black!75},
    edge label/.style={font=\scriptsize, fill=white, inner sep=1pt}
]

\def\yA{2.7}
\def\yB{0.9}
\def\yC{-0.9}
\def\yD{-2.7}

\node[font=\small\bfseries, text=black!75] at (3.4,  4.7) {1. Tokenisation};
\node[font=\small\bfseries, text=black!75] at (6.3,  4.7) {2. Hilbert routing};
\node[font=\small\bfseries, text=black!75] at (9.2,  4.7) {3. Distance-aware scan};
\node[font=\small\bfseries, text=black!75] at (14.6, 4.7) {4. Fusion \& output};

\node[input node] (t_in)   at (0,  3.6) {Flow time\\$t$};
\node[input node] (coords) at (0,  1.8) {Coordinates\\$\mathbf{X}_C, \mathbf{X}_Q$};
\node[input node] (y_ctx)  at (0,  0.0) {Context\\$\mathbf{Y}_C$};
\node[input node] (f_qry)  at (0, -1.8) {Noisy queries\\$\mathbf{Y}_t$};

\node[et back, minimum width=2.4cm, minimum height=6.7cm, anchor=north] (etok) at (3.6, 4.2) {};
\node[font=\footnotesize\bfseries, text=myred!80!black, align=center] at (3.6, 3.6)
    {Time embedding\\$\mathbf{e}_t$};
\node[process block, minimum width=2.4cm, minimum height=6.0cm] (tok) at (3.2, 0)
    {Tokenise\\$\mathbf{E}\!\in\!\mathbb{R}^{(N+M)\times d}$};

\draw[conn] (coords.east) -- (tok.west |- coords.east);
\draw[conn] (y_ctx.east)  -- (tok.west |- y_ctx.east);
\draw[conn] (f_qry.east)  -- (tok.west |- f_qry.east);
\draw[conn, color=myred!80!black] (t_in.east) -- (2.4, 3.6);

\node[sort block] (sA) at (6.3, \yA) {Sort $\pi_1$};
\node[sort block] (sB) at (6.3, \yB) {Sort $\pi_1^{\mathrm{rev}}$};
\node[sort block] (sC) at (6.3, \yC) {Sort $\pi_2$};
\node[sort block] (sD) at (6.3, \yD) {Sort $\pi_2^{\mathrm{rev}}$};

\draw[conn] (etok.east |- sA) -- (sA.west);
\draw[conn] (etok.east |- sB) -- (sB.west);
\draw[conn] (etok.east |- sC) -- (sC.west);
\draw[conn] (etok.east |- sD) -- (sD.west);

\node[mamba back] (mA_b) at (9.40, 3.10) {};
\node[mamba mid]  (mA_m) at (9.20, 2.90) {};
\node[mamba front] (mA)  at (9.0,  \yA)  {Mamba-2 $(\times L)$};
\node[mamba back] (mB_b) at (9.40, 1.30) {};
\node[mamba mid]  (mB_m) at (9.20, 1.10) {};
\node[mamba front] (mB)  at (9.0,  \yB)  {Mamba-2 $(\times L)$};
\node[mamba back] (mC_b) at (9.40, -0.50) {};
\node[mamba mid]  (mC_m) at (9.20, -0.70) {};
\node[mamba front] (mC)  at (9.0,  \yC)   {Mamba-2 $(\times L)$};
\node[mamba back] (mD_b) at (9.40, -2.30) {};
\node[mamba mid]  (mD_m) at (9.20, -2.50) {};
\node[mamba front] (mD)  at (9.0,  \yD)   {Mamba-2 $(\times L)$};

\draw[conn] (sA) -- (mA) node[edge label, midway, text=myteal!80!black] {$\delta_d$};
\draw[conn] (sB) -- (mB) node[edge label, midway, text=myteal!80!black] {$\delta_d$};
\draw[conn] (sC) -- (mC) node[edge label, midway, text=myteal!80!black] {$\delta_d$};
\draw[conn] (sD) -- (mD) node[edge label, midway, text=myteal!80!black] {$\delta_d$};

\node[sort block] (uA) at (12.1, \yA) {Unsort};
\node[sort block] (uB) at (12.1, \yB) {Unsort};
\node[sort block] (uC) at (12.1, \yC) {Unsort};
\node[sort block] (uD) at (12.1, \yD) {Unsort};

\draw[conn] (mA) -- (uA);
\draw[conn] (mB) -- (uB);
\draw[conn] (mC) -- (uC);
\draw[conn] (mD) -- (uD);

\node[fuse block, minimum width=1.7cm, minimum height=6.0cm] (fuse) at (14.6, 0)
    {Linear\\merge};

\draw[conn] (uA.east) -- (fuse.west |- uA.east);
\draw[conn] (uB.east) -- (fuse.west |- uB.east);
\draw[conn] (uC.east) -- (fuse.west |- uC.east);
\draw[conn] (uD.east) -- (fuse.west |- uD.east);

\node[process block, minimum width=2.0cm, minimum height=1.2cm] (readout) at (17.0, 0)
    {Output\\projection};
\node[input node, font=\small\bfseries] (out) at (20.0, 0)
    {$v_\theta(\mathbf{Y}_t, t, \mathbf{c})$};

\draw[conn] (fuse.east) -- (readout.west);
\draw[conn] (readout.east) -- (out.west);

\end{tikzpicture}}
\caption{MFNP parametrises the velocity field $v_\theta(\mathbf{Y}_t, t, \mathbf{c})$ of \S\ref{sec:cfm} in four stages. The stages are tokenisation of context values and the noisy query slice $\mathbf{Y}_t$ with frozen Random Fourier Feature (RFF) coordinate embeddings; multi-directional Hilbert routing under $\Pi = \{\pi_1, \pi_1^{\mathrm{rev}}, \pi_2, \pi_2^{\mathrm{rev}}\}$; distance-aware Mamba-2 scans modulated by an adaLN-Zero flow-time embedding $\mathbf{e}_t$; and inverse-permutation fusion to per-query velocities. Trained with $\mathcal{L}_{\mathrm{CFM}}$, reflow-distilled to a one-step Euler student, and CRPS fine-tuned. Full pipeline in Appendix~\ref{app:detailed-algs}, Algorithm~\ref{alg:MFNP_detailed}.}
\label{fig:arch}
\end{figure}

\textbf{MambaFlowNP} (MFNP) avoids this compression by maintaining
a hidden state at every token position throughout a
multi-directional state-space scan (Figure~\ref{fig:arch}). Context
and query tokens share a single sequence, and the model predicts a
flow-matching~\citep{lipman2022flow,liu2022flow} velocity at each
query location. A single joint-predictive draw for all the query
locations requires batched ODE integration of the velocity, which
typically entails tens of network function evaluations (NFE)
(Table~\ref{tab:distillation}). MFNP mitigates this cost via a single
round of reflow distillation~\citep{liu2022flow} to enable inference
with only one NFE. As in latent-variable NPs, dependence between queries is
carried by a shared latent, here a random-function initial condition
transported by the learned flow (\S\ref{sec:consistency}).

\paragraph{Contributions.}
\begin{enumerate}[leftmargin=2em]
\item We introduce MambaFlowNP (MFNP), a joint-predictive
    NP that processes observations and prediction locations
    together in a shared sequence, scanned in multiple directions.
    Its computational cost grows linearly with the total number
    of points.

    \item Through controlled ablations, we identify and
        quantify a single-state compression bottleneck in
        state-space NPs. A variant that compresses the context into one terminal
        hidden state has $10$--$17\%$ higher MSE than the full model on
        synthetic tasks and $58$--$65\%$ higher MSE on NOAA. This finding
        motivates MFNP's use of hidden states at individual query
        positions.

    \item Across three NOAA temperature interpolation tiers,
        MFNP achieves lower RMSE and CRPS than the strongest
        joint-predictive baselines, with sampling speedups of
        $114$--$784{\times}$. We separately assess the contributions
        of reflow distillation and the linear-time scan to these
        speedups.

    \item We demonstrate computational scalability to
        $10^{6}$ points, with a single forward pass completing in
        under a second on one NVIDIA L40S GPU. Each forward pass
        requires $O(K)$ compute, where $K$ is the total number of
        context and query points. All tested quadratic-time
        baselines exceed GPU memory at this scale.
\end{enumerate}

\begin{figure}[!htbp]
\centering
\includegraphics[width=\linewidth]{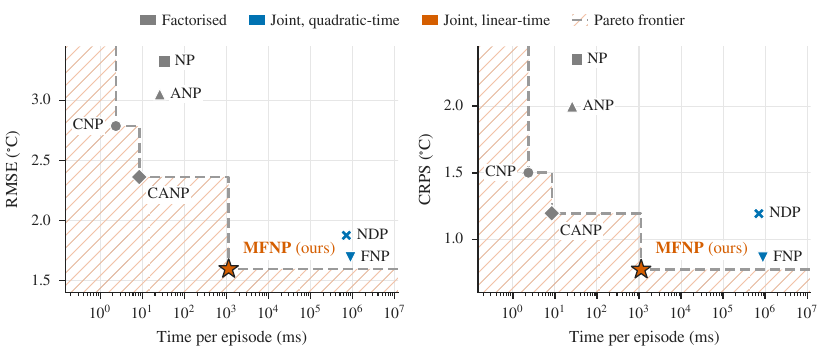}
\caption{Per-episode latency vs.\ RMSE (left) and CRPS (right) on continental US NOAA ($K{=}5{,}580$). Means $\pm 1$ SE over $10$ seeds at $N_{\text{samples}}{=}128$; the hatched staircase is the empirical Pareto frontier. TNPA exceeds the per-cell budget (\S\ref{sec:protocol}); Utah and California frontiers in Appendix~\ref{app:full-noaa}.}
\label{fig:noaa-pareto}
\end{figure}

\section{Background}

\subsection{Neural processes}
\label{sec:np}

A neural process defines a conditional distribution $p(\mathbf{Y}_Q \mid \mathbf{X}_C, \mathbf{Y}_C, \mathbf{X}_Q)$ over target values $\mathbf{Y}_Q \in \mathbb{R}^{M}$ at query locations $\mathbf{X}_Q \in \mathbb{R}^{M \times D}$, given a context of $N$ input--output pairs $\mathbf{X}_C \in \mathbb{R}^{N \times D}$, $\mathbf{Y}_C \in \mathbb{R}^{N}$. Let $K = N + M$ denote the total number of points. We tokenise each point separately, so $K$ is also the sequence length.
Training is episodic. Each iteration samples a dataset from a stochastic process, partitions points into context and query, and fits the predictive distribution against the ground-truth query values.
At test time, the context and query locations of a new dataset are passed through the trained model in a single forward pass. NPs do not specify an explicit prior and then apply Bayes' rule; instead, they learn an implicit prior over functions from the data-generating stochastic process during episodic training, and the forward pass through $(\mathbf{X}_C, \mathbf{Y}_C)$ acts as an amortised update producing an approximation to the posterior predictive $p(\mathbf{Y}_Q \mid \mathbf{X}_C, \mathbf{Y}_C, \mathbf{X}_Q)$ over the query locations. We use \emph{predictive} and \emph{posterior predictive} interchangeably for this distribution.

\subsection{Conditional flow matching and reflow distillation}
\label{sec:cfm}
Flow matching~\citep{lipman2022flow, albergo2023building,liu2022flow}
models approximate a velocity field $u_t$ that transports samples
from a source distribution $p_0$ to the conditional data distribution
$p_{1}(\mathbf{Y}_1 \mid \mathbf{c})\equiv p_{\mathrm{data}}
(\mathbf{Y}_1 \mid \mathbf{c})$, where $\mathbf{c} = \{\mathbf{X}_C,
\mathbf{Y}_C, \mathbf{X}_Q\}$ denotes the neural process conditioning.
Here $\mathbf{Y}_1$ equals the query values $\mathbf{Y}_Q\in\mathbb{R}^M$
targeted by the predictive, so $\mathbf{Y}_t\in\mathbb{R}^M$ is indexed
by the query locations. The velocity field is determined by a choice of conditional flow path
or interpolant $\mathbf{Y}_{t}=a_t\mathbf{Y}_1+b_t\mathbf{Y}_0$,
\begin{equation}
    u_{t}(\mathbf{Y}_t; \mathbf{c})=\int\frac{d\mathbf{Y}_t}{dt}
    \frac{p_t(\mathbf{Y}_t \mid \mathbf{Y}_1, \mathbf{c})
    p_{1}(\mathbf{Y}_1 \mid \mathbf{c})}{p_t(\mathbf{Y}_t \mid \mathbf{c})}
    d\mathbf{Y}_1
\end{equation}
where $a_t, b_t\in\mathbb{R}$, $\mathbf{Y}_1\sim p_1(\cdot \mid
\mathbf{c})$, $\mathbf{Y}_0\sim p_0$, and $\frac{d\mathbf{Y}_t}{dt}$ is the
\emph{conditional} velocity. Directly regressing a neural
parametrisation $v_{\theta}(\mathbf{Y}_t, t, \mathbf{c})$ to
$u_t(\mathbf{Y}_t;\mathbf{c})$ is intractable due to the marginalisation
in the definition of $u_t$. The \emph{conditional} flow-matching loss is tractable and has the same gradient
\citep{lipman2022flow},
\begin{equation}
    \mathcal{L}_{\mathrm{CFM}}(\theta)=\mathbb{E}_{t,\mathbf{Y}_1,
    \mathbf{Y}_0,\mathbf{c}}\left[\bigg\lVert v_{\theta}
    (\mathbf{Y}_t, t, \mathbf{c}) -\frac{d\mathbf{Y}_t}{dt}
    \bigg\rVert^2 \right]
\end{equation}
with $t\sim\mathcal{U}(0,1)$.
Samples are drawn by integrating $\hat{\mathbf{Y}}_1=
\mathbf{Y}_0+\int_{0}^{1}v_{\theta}(\mathbf{Y}_t, t, \mathbf{c})dt$
numerically (e.g.\ Euler). 
Drawing inspiration from work on data-dependent source distributions \citep{tong2024improving,kollovieh2025flow}, we use a query-geometry-dependent Gaussian source distribution $p_0(\Y_0\mid \mathbf{X}_Q)=\mathcal{N}(0,\Sigma(\mathbf{X}_Q))$, obtained by evaluating a random-feature Gaussian-process draw at the query locations. This gives a spatially correlated source, of rank at most $F$, that shortens the transformation from source samples to smooth spatial fields (\S\ref{sec:sampling}). The network also receives a per-draw latent $\mathbf{z}$ and routing offset $\mathbf{o}$ (\S\ref{sec:method}), which we omit from the notation $v_\theta(\mathbf{Y}_t,t,\mathbf{c})$. The gradient equivalence above assumes a network without these inputs. With them, the minimiser of $\mathcal{L}_{\mathrm{CFM}}$ is $\mathbb{E}[\mathbf{Y}_1-\mathbf{Y}_0\mid\mathbf{Y}_t,t,\mathbf{c},\mathbf{z},\mathbf{o}]$ rather than $u_t$; reflow and CRPS fine-tuning (\S\ref{sec:crps}) then fit the one-step map directly. We use the
linear interpolant $a_t=t$, $b_t=(1-t)$, with conditional velocity
$\frac{d\mathbf{Y}_t}{dt}=\mathbf{Y}_1-\mathbf{Y}_0$. Although this
conditional flow has straight trajectories, the underlying velocity field typically induces curved
trajectories, leading to inaccurate results with few solver steps \citep{liu2022flow,geng2026mean}. To obtain high-quality samples with fewer steps, we use reflow distillation~\citep{liu2022flow}. A pre-trained teacher model generates
coupled $(\Y_0,\hat{\Y}_1)$ pairs, and the student model is optimised on $\mathcal{L}_{\mathrm{CFM}}$ against these couplings rather than independently drawn source and target samples. The full training pipeline is described in \S\ref{sec:training} and Appendix~\ref{app:distillation}.

\section{Related work}

\paragraph{Factorised and latent-variable NPs.}
Conditional Neural Processes~\citep{garnelo2018conditional}
encode the context with a permutation-invariant set encoder and
predict independent Gaussian distributions at query locations.
Latent-variable NPs~\citep{garnelo2018neural}
introduce dependence across queries through a shared
Gaussian latent, trained under a variational ELBO.
Attentive NP~\citep{kim2019attentive} replaces mean-pooling with
cross-attention from each query to the context.
Convolutional Conditional
NPs~\citep{gordon2019convolutional} use grid-based convolutions
to encode translation equivariance, an inductive bias
suited to stationary spatial fields.
AR-CNP~\citep{bruinsma2023arcnp}
introduces cross-query dependence by sampling from a CNP
autoregressively, requiring $M$ sequential forward passes
for $M$ queries.

\paragraph{Joint-predictive NPs.}
Gaussian NPs~\citep{bruinsma2021gaussian} explicitly model
cross-query correlations through a joint Gaussian predictive.
TNPA~\citep{nguyen2022transformer} decodes queries autoregressively
under causal self-attention.
NDP~\citep{dutordoir2023neural} and
FNP~\citep{abuhamad2025flownp} use transformer backbones
trained with score-based diffusion~\citep{song2021scorebased}
and flow-matching~\citep{lipman2022flow} objectives, respectively.
In these transformer models, self-attention incurs
$O(K^2)$ compute per network evaluation.
Spatial coordinates enter through token embeddings,
whereas MFNP also uses them to determine the scan order.

\paragraph{Mamba on point sets.}
Mamba~\citep{gu2023mamba} and Mamba-2~\citep{dao2024mamba2}
introduced selective state-space models with input-dependent
gating and computational cost linear in sequence length.
VMamba~\citep{liu2024vmamba} scans regular image grids in four
directions. MFNP adapts this four-direction scheme to irregular point sets
through two Hilbert orderings and their reversals.
PointMamba~\citep{liang2024pointmamba} serialises
3D point clouds along Hilbert curves for supervised shape
analysis, while Voxel Mamba~\citep{zhang2024voxelmamba}
applies Hilbert serialisation to sparse voxels for 3D detection.
These methods use spatial scans for visual recognition tasks.
MFNP applies multi-directional scans to amortised
joint-predictive inference for continuous 2D fields,
with episodic training over varying numbers of context
and query points $(N,M)$.

\paragraph{Scalable GPs and spatial statistics.}
Exact GP inference for kriging requires an $O(N^3)$ covariance
factorisation for $N$ observed locations.
Sparse variational GPs~\citep{titsias2009variational},
Vecchia approximations and nearest-neighbour
GPs~\citep{vecchia1988estimation, datta2016hierarchical},
and structured kernel interpolation~\citep{wilson2015kernel}
reduce this computational burden, and the SPDE approach represents Mat\'ern fields as sparse
Gaussian Markov random fields~\citep{lindgren2011explicit}. These approaches make
inference within an individual field more tractable, but each still fits
hyperparameters per field.
NPs amortise inference across fields. Episodic training
learns a map from context observations to a predictive
distribution, which can then be applied to a new field without
further optimisation.

\paragraph{In-context learning and prior-fitted networks.}
NPs learn to infer from context observations without updating
their parameters at test time.
Prior-fitted networks~\citep{muller2022transformers}
similarly train transformers on datasets sampled from
a prior to approximate the posterior predictive.
Mamba has also been shown to learn regression tasks
in context~\citep{grazzi2024mamba}.

MFNP builds on established components, namely Mamba-2
scans~\citep{dao2024mamba2}, Hilbert
serialisation~\citep{liang2024pointmamba, zhang2024voxelmamba},
conditional flow matching~\citep{lipman2022flow}, GP-based source distributions~\citep{kollovieh2025flow}, and reflow
distillation~\citep{liu2022flow}. Our contributions are the shared
context--query scan for joint-predictive neural processes and
the analysis of how scan routing affects predictive performance.

\section{MambaFlowNP}
\label{sec:method}
\label{sec:theory}

MFNP processes irregularly placed observations in one or two dimensions with an $O(K)$
neural recurrence. A direct SSM-based NP adaptation compresses the
context into a single fixed-dimensional hidden state and conditions
a query decoder on that state. MFNP avoids the compression by
maintaining a hidden state at every token position throughout the
scan. Context and query tokens share the sequence, so each
query is decoded from its own position. The single-state ablation
forces the context through one hidden vector before reading queries
(\S\ref{sec:ablation-results}).

\subsection{Architecture}

For each joint draw, a shared latent $\mathbf{z}$ determines
the random initial values at the query locations and conditions
the network. Given the current query values $\mathbf{Y}_t$, flow
time $t$ and context $\mathbf{c}$, the network computes the velocity
$v_\theta(\mathbf{Y}_t,t,\mathbf{c})$ in the four stages shown in
Figure~\ref{fig:arch}.

\textbf{(1) Tokenise.} For each point, we concatenate a
coordinate embedding computed using
frozen Random Fourier Features~\citep{rahimi2007random}
with its scalar value, which is the observed value at a context point
or $Y_{t,i}$ at a query. Context and query tokens use separate
linear projections.

\textbf{(2) Route.} We use Hilbert
curves~\citep{sagan1994sfc, butz1971hilbert}
to order the combined context and query points so that nearby
locations tend to remain close in the sequence. Two orderings,
$\pi_1$ and $\pi_2$, use the original and swapped coordinate axes.
Together with their reversals, they form
\[
\Pi = \{\pi_1, \pi_1^{\mathrm{rev}}, \pi_2, \pi_2^{\mathrm{rev}}\}.
\]
A shared random offset $\mathbf{o}$ shifts the routing grid
to reduce artefacts at fixed curve boundaries. In one dimension,
we scan points in ascending and descending coordinate order, repeating these two orderings to fill the four direction slots (Appendix~\ref{app:detailed-algs}).

\textbf{(3) Scan.} A Mamba-2 block
propagates information through a recurrent hidden state.
Context tokens write to the recurrent state. Query tokens
read the state at their own positions but do not write to it, since they enter
the scan with zero input and zero step size; the block has no short convolution.
A single $L$-layer stack of Mamba-2 blocks, with parameters
shared across directions, processes each direction-permuted
sequence. Every block is conditioned through
adaLN-Zero~\citep{peebles2023scalable} on the flow time and the
shared latent $\mathbf{z}$. Its state-update gate also uses the
distance $\delta_{d,i}$ to the nearest preceding context point in
the scan order.

\textbf{(4) Fuse and read out.} The per-direction
hidden states are restored to the original point order and
linearly fused, and a final adaLN-conditioned projection maps
each query's fused state to its velocity. Implementation settings
and the full forward pass are given in
Appendix~\ref{app:detailed-algs}, Algorithm~\ref{alg:MFNP_detailed}.

\subsection{Training and sampling}
\label{sec:sampling}
\label{sec:crps}

\textbf{Random-function source distribution.}
Each draw samples a latent $\mathbf{z} \sim \mathcal{N}(0,I_F)$
and sets
\[
\text{}Y_{0,i}
= \boldsymbol{\psi}(\mathbf{x}_{Q,i})^{\top}\mathbf{z},
\]
where $\boldsymbol{\psi}$ is a fixed $F$-dimensional random Fourier feature map.
This is the source $p_0$ of \S\ref{sec:cfm}. It evaluates the same random function at every query location,
providing spatially dependent initial values.
It replaces the standard $\mathbf{Y}_0$ draw in the CFM loss and at
inference (Algorithms~\ref{alg:MFNP_train_detailed}
and~\ref{alg:MFNP_inference_detailed}).

\textbf{Flow matching and distillation.}
We first train a teacher using the conditional flow-matching
objective in \S\ref{sec:cfm}. The teacher then generates paired
initial and final samples for one round of reflow distillation.
The resulting student generates a joint draw with one Euler step,
requiring a single network evaluation
(Appendix~\ref{app:distillation}).

\textbf{CRPS fine-tuning.}
On NOAA, a short final stage fine-tunes the student on the continuous ranked
probability score of its own one-step samples. For each training
episode, we draw $S$ samples $\hat{y}^{(1)}, \dots, \hat{y}^{(S)}$ at
every query and minimise the fair CRPS~\citep{gneiting2007strictly,ferro2014fair}
\begin{equation}
    \mathcal{L}_{\mathrm{CRPS}}
    = \frac{1}{S}\sum_s \bigl|\hat{y}^{(s)}-y\bigr|
    - \frac{1}{2S(S-1)}\sum_{s\neq s'}
      \bigl|\hat{y}^{(s)}-\hat{y}^{(s')}\bigr|,
\end{equation}
averaged over queries. The stage early-stops on the CRPS of
held-out validation episodes (Appendix~\ref{app:hparams}). Baselines do not receive this stage.

\subsection{Properties and computational cost}
\label{sec:consistency}

For distinct Hilbert sort keys, deterministic
sorting makes MFNP invariant to context-row permutations and
equivariant to query-row permutations.

It is also consistent under query-set marginalisation.
Queries only read the scan state, and all randomness shared
between queries is the latent $\mathbf{z}$ and offset $\mathbf{o}$
of each draw, so each draw is one random function of the context
evaluated at the queries. The joint predictive over any query set therefore has
at most $F$ degrees of freedom, as in latent-variable NPs. For the same context, latent and offset,
adding, removing, or chunking queries leaves the values at the
other locations unchanged, and, given the context, the predictive defines a
Kolmogorov-consistent stochastic process over query locations. Recent transformer NPs
relax this property for efficiency~\citep{hassan2025efficient,
mortimer2026incremental}. Because queries do not write to the scan
state, the context states can be computed once and reused across
query sets for a fixed draw and flow time.

\begin{figure}[t]
\centering
\includegraphics[width=0.58\linewidth]{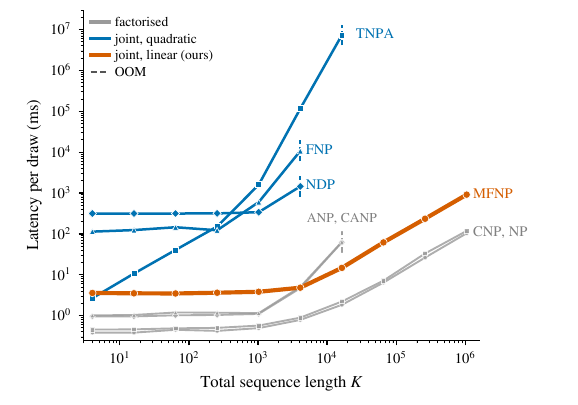}
\caption{End-to-end sampling latency vs.\ $K$ (log--log) on 2D Gaussian Mat\'ern-$\nicefrac{3}{2}$, batch$=1$, $N_{\text{samples}}{=}1$ (1-NFE Euler for MFNP; published step count per baseline). Colour gives the model class; a short dashed line through a curve's last point means the next $K$ runs out of memory. Numbers in Table~\ref{tab:scaling}.}
\label{fig:scaling}
\end{figure}

\textbf{Inference complexity.}
The random-function source costs $O(MF)$, which is linear in
the query count for fixed feature dimension $F$.
The Mamba-2 recurrence has computational complexity $O(K)$ in
sequence length by construction. The routing permutations are
obtained by sorting Hilbert keys, also $O(K)$
in the operating regime (Appendix~\ref{app:sort-scaling}). Each forward pass therefore costs $O(K)$ for fixed feature dimension, state size and depth.

\section{Experimental setup}
\label{sec:experiments}

\subsection{Tasks}
\label{sec:tasks}

\paragraph{Synthetic Gaussian processes.}
Each episode draws from a fresh GP, following the standard settings of TNP~\citep{nguyen2022transformer}. 1D tasks sample $K \le
K_{\max}{=}128$ points uniformly in $[-2,2]$; 2D tasks sample
$K \le K_{\max}{=}256$ points uniformly in $[0,1]^2$. The simulator
draws a latent field $f \sim \mathcal{GP}(0, k_{\ell,\sigma_f^2})$
and observations $y = f + \varepsilon$, $\varepsilon \sim
\mathcal{N}(0, \sigma_n^2 I)$. Kernel hyperparameters are resampled
per episode from $\ell \sim \mathcal{U}(0.1, 0.6)$ and $\sigma_f \sim \mathcal{U}(0.1, 1.0)$,
with $\sigma_n = 0.02$, so the model must infer smoothness and amplitude
from context. Context and query sizes vary per episode
with $N \sim \mathcal{U}\{16, K_{\max} - 16\}$ and
$M \sim \mathcal{U}\{16, K_{\max} - N\}$. We use RBF, Mat\'ern-
$\nicefrac{3}{2}$, and Mat\'ern-$\nicefrac{5}{2}$ kernels in 1D,
and Mat\'ern-$\nicefrac{3}{2}$ in 2D.

\begin{figure}[t]
\centering
\includegraphics[width=0.7\linewidth]{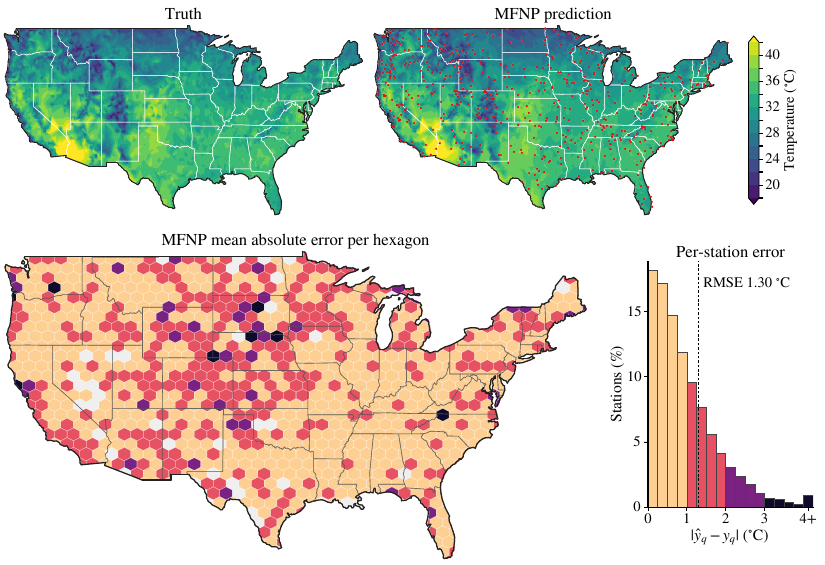}
\caption{Continental US ($K{=}5{,}580$) posterior on a held-out date, $10\%/90\%$ context/query split. Top, truth and MFNP posterior mean (red dots are context stations), interpolated for display. Bottom, MFNP absolute error averaged over the query stations in each ${\sim}110$\,km hexagon (grey hexagons hold none), with the per-station histogram as legend and the RMSE dashed. Population-averaged accuracy in Figure~\ref{fig:noaa-ranges}.}
\label{fig:noaa-qual-us}
\end{figure}

\paragraph{NOAA temperature interpolation.}
We use daily maximum temperature (TMAX) records from the NOAA Global
Historical Climatology Network~\citep{menne2012overview,menne2012ghcnd}
over 2005-01-01 to 2024-12-31, split temporally into train
($\le$~2016), validation (2017--2019), and test (2020--2024) sets.
These cover \textbf{Utah}
($K{=}222$), \textbf{California} ($K{=}486$) and the continental
\textbf{US} ($K{=}5{,}580$), three geographic tiers of increasing extent. The same station set is reused across
splits, so generalisation is across dates rather than locations; the task is therefore infilling at stations seen during training.
Each NOAA episode draws a date uniformly from the appropriate split and
partitions that date's observed stations $50/50$ at random into
context and query. Coordinate projection, station selection, and
target scaling are in Appendix~\ref{app:hparams}.

\begin{table}[t]
\centering
\caption{1D and 2D Gaussian Mat\'ern-$\nicefrac{3}{2}$ ($K{=}128$, $256$). MSE, CRPS ($\downarrow$); C2ST (target $0.5$); $95\%$ coverage (target nominal); time = per-episode ms, MFNP and baselines on one NVIDIA L40S. Mean $\pm$ SE over 10 evaluation seeds; bold = best per column (within 1 SE). \textsuperscript{\dag} our method; full kernel sweep in Appendix~\ref{app:synth-full}.}
\label{tab:synth-main}
\scriptsize
\setlength{\tabcolsep}{4pt}
\begin{tabular}{@{} l c c c c c @{}}
\toprule
 & \multicolumn{2}{c}{\textbf{Predictive Quality}} & \multicolumn{2}{c}{\textbf{Calibration Quality}} & \\
\cmidrule(lr){2-3} \cmidrule(lr){4-5}
\textbf{Model} & \textbf{MSE}\,$\downarrow$ & \textbf{CRPS}\,$\downarrow$ & \textbf{C2ST}\,$\rightarrow\!0.5$ & \textbf{Cov-95}\,$\rightarrow\!0.95$ & \textbf{Time\,(ms)}\,$\downarrow$ \\
\midrule

\multicolumn{6}{l}{\textit{1D Gaussian -- Mat\'ern-$\nicefrac{3}{2}$} ($K{=}128$)} \\[2pt]

\multicolumn{6}{@{}l}{\;\; \textit{Joint predictives, quadratic-time}} \\
  TNPA & $\mathbf{0.0110\pm0.0003}$ & $\mathbf{0.0367\pm0.0004}$ & $\mathbf{0.499\pm0.003}$ & $\mathbf{0.946\pm0.001}$ & $735.7\pm0.4$ \\
  NDP  & $0.1720\pm0.0028$ & $0.2508\pm0.0014$ & $0.688\pm0.018$ & $0.996\pm0.000$ & $(2.174\pm0.004){\times}10^{3}$ \\
  FNP  & $\mathbf{0.0107\pm0.0002}$ & $0.0398\pm0.0004$ & $0.532\pm0.011$ & $0.752\pm0.001$ & $(1.253\pm0.001){\times}10^{3}$ \\
\multicolumn{6}{@{}l}{\;\; \textit{Joint predictives, linear-time}} \\
  MFNP\textsuperscript{\dag} & $0.0111\pm0.0003$ & $0.0398\pm0.0004$ & $0.537\pm0.006$ & $0.938\pm0.001$ & $\mathbf{29.5\pm0.2}$ \\
\midrule

\multicolumn{6}{l}{\textit{2D Gaussian -- Mat\'ern-$\nicefrac{3}{2}$} ($K{=}256$)} \\[2pt]

\multicolumn{6}{@{}l}{\;\; \textit{Joint predictives, quadratic-time}} \\
  TNPA & $\mathbf{0.0126\pm0.0004}$ & $\mathbf{0.0453\pm0.0006}$ & $\mathbf{0.488\pm0.002}$ & $\mathbf{0.938\pm0.001}$ & $(3.480\pm0.002){\times}10^{3}$ \\
  NDP  & $0.0443\pm0.0007$ & $0.1089\pm0.0007$ & $0.668\pm0.009$ & $0.992\pm0.000$ & $(3.158\pm0.004){\times}10^{3}$ \\
  FNP  & $0.0137\pm0.0004$ & $0.0503\pm0.0007$ & $0.573\pm0.019$ & $0.630\pm0.001$ & $(3.526\pm0.002){\times}10^{3}$ \\
\multicolumn{6}{@{}l}{\;\; \textit{Joint predictives, linear-time}} \\
  MFNP\textsuperscript{\dag} & $0.0167\pm0.0005$ & $0.0537\pm0.0005$ & $0.554\pm0.005$ & $0.971\pm0.000$ & $\mathbf{103.6\pm0.3}$ \\
\bottomrule
\end{tabular}
\end{table}

\subsection{Training}
\label{sec:training}

MFNP trains a CFM teacher for $80$k steps and
a reflow student for a further $20$k steps. On NOAA, this is followed by the CRPS
fine-tuning stage of \S\ref{sec:crps}. On Utah, California and the US, the
student trains on $5{,}000$, $5{,}000$ and $1{,}000$ teacher couplings. CRPS
fine-tuning uses batches of $8$, $8$ and $2$ episodes with $8$, $8$ and $4$
samples per episode, a learning rate of $10^{-4}$ scaled by the square
root of the batch size relative to $8$; it is scored on $64$
validation episodes from 2017--2019 every $1{,}000$ steps, stops after
$5$ evaluations without improvement or at $40$k steps, and keeps the
best weights. Baselines train for
$100$k steps with no student or CRPS phase; on NOAA, MFNP receives up to $140$k optimiser updates in total. Per-cell hyperparameters (hidden size, depth, learning rate, batch size)
come from Bayesian-optimisation sweeps on validation
MSE, with a single training run at the BO-best configuration.
Optimiser and schedule are shared across models (AdamW,
warmup-cosine to $10^{-5}$, EMA $\beta{=}0.9995$); per-cell
values and the BO protocol are in Appendix~\ref{app:hparams},
and the reflow coupling and solver details are in Appendix~\ref{app:distillation}.
All training and evaluation run on NVIDIA L40S GPUs.
Each (model, task) cell is capped at $100$ GPU-hours covering
training and all $10$ evaluation seeds; cells exceeding the cap
are reported as \textsc{t/o}.

\subsection{Evaluation protocol}
\label{sec:protocol}

\textbf{Metrics.} Predictive quality is scored by mean-squared error (MSE) on the
predictive mean and the continuous ranked probability score
(CRPS)~\citep{gneiting2007strictly}, each averaged over
$N_{\text{samples}}{=}128$ draws per target shared across methods
to control the $O(1/N_{\text{samples}})$ bias in the standard empirical CRPS estimator.
Calibration is scored by the classifier two-sample test
(C2ST)~\citep{lopezpaz2017revisiting} against GP reference draws
(synthetic only) and the empirical coverage of the central $95\%$ interval of the $128$ samples,
both computed at each query location and averaged over locations. All numbers are
mean $\pm$ one standard error across $10$ evaluation seeds for a single trained model per cell, each
covering $256$ synthetic episodes or $1{,}024$ NOAA episodes on a
fresh date; evaluation sets are cached so every method sees
identical test data.

\textbf{Latency.} Per-episode wall-clock is the median of $10$
timed forward passes after $5$ warm-ups on a single NVIDIA L40S,
each pass a full $N_{\text{samples}}{=}128$-draw call on one
held-out episode, that is, the time to generate $128$ joint
predictive draws over all query locations. Timings include the
Hilbert sort. MFNP times use the release sampler, $8$ routing offsets of $16$ draws each. Each method runs its deployment configuration.
MFNP uses 1-NFE Euler from the reflow student, FNP and NDP use
their published ODE/diffusion step counts, and TNPA uses autoregressive
decoding. TNPA at NOAA US ($K{=}5{,}580$) is reported as \textsc{t/o}
because its serial-within-a-draw decoding exceeds the per-cell
budget by an order of magnitude (Appendix~\ref{app:hparams}).

\section{Results}
\label{sec:results}

\subsection{Synthetic Gaussian processes}
\label{sec:synthetic}

MFNP is competitive with the strongest joint-predictive baselines
on predictive quality at matched point counts and is
$25$--$42{\times}$ faster (Table~\ref{tab:synth-main}; full kernel
sweep in Appendix~\ref{app:synth-full}). On 1D
Mat\'ern-$\nicefrac{3}{2}$, MFNP matches TNPA on MSE within one SE
and FNP on CRPS, at $30$\,ms per episode against $736$\,ms for TNPA. On 2D
Mat\'ern-$\nicefrac{3}{2}$, MFNP records ${\sim}1.2{\times}$ higher MSE than FNP at
${\sim}34{\times}$ lower latency. MFNP coverage stays within three points of $0.95$ on the Mat\'ern
kernels and nine points on RBF, whereas FNP under-covers by $20$--$32$ points and NDP over-covers.

\begin{table}[b]
\centering
\caption{NOAA temperature interpolation on Utah ($K{=}222$), California ($K{=}486$), and the continental US ($K{=}5{,}580$); joint-predictive baselines only (full table with factorised baselines in Table~\ref{tab:noaa-full}). MSE (\textdegree C)$^2$, RMSE, CRPS \textdegree C ($\downarrow$); Cov-95 (target $0.95$); C2ST omitted (no tractable reference posterior on real fields); time = per-episode ms on one NVIDIA L40S. Mean $\pm$ SE over 10 evaluation seeds, denormalised scale; bold = best per column within tier (within 1 SE). \textsuperscript{\dag} our method.}
\label{tab:noaa-main}
\scriptsize
\setlength{\tabcolsep}{4pt}
\begin{tabular}{@{} l c c c c c @{}}
\toprule
 & \multicolumn{3}{c}{\textbf{Predictive Quality}} & \textbf{Calibration} & \\
\cmidrule(lr){2-4} \cmidrule(lr){5-5}
\textbf{Model} & \textbf{MSE (\textdegree C)$^2$\,$\downarrow$} & \textbf{RMSE (\textdegree C)\,$\downarrow$} & \textbf{CRPS (\textdegree C)\,$\downarrow$} & \textbf{Cov-95}\,$\rightarrow\!0.95$ & \textbf{Time\,(ms)}\,$\downarrow$ \\
\midrule

\multicolumn{6}{l}{\textit{NOAA Utah} ($K{=}222$)} \\[2pt]
\multicolumn{6}{@{}l}{\;\; \textit{Joint predictives, quadratic-time}} \\
  TNPA & $2.524\pm0.007$ & $1.589\pm0.002$ & $0.783\pm0.001$ & $0.846\pm0.000$ & $(4.923\pm0.004){\times}10^{3}$ \\
  NDP  & $4.665\pm0.012$ & $2.160\pm0.003$ & $1.221\pm0.002$ & $0.742\pm0.000$ & $(2.414\pm0.004){\times}10^{3}$ \\
  FNP  & $2.712\pm0.007$ & $1.647\pm0.002$ & $0.853\pm0.001$ & $0.730\pm0.000$ & $(3.071\pm0.003){\times}10^{3}$ \\
\multicolumn{6}{@{}l}{\;\; \textit{Joint predictives, linear-time}} \\
  MFNP\textsuperscript{\dag} & $\mathbf{2.464\pm0.016}$ & $\mathbf{1.570\pm0.005}$ & $\mathbf{0.767\pm0.002}$ & $\mathbf{0.940\pm0.000}$ & $\mathbf{43.3\pm0.4}$ \\
\addlinespace

\midrule

\multicolumn{6}{l}{\textit{NOAA California} ($K{=}486$)} \\[2pt]
\multicolumn{6}{@{}l}{\;\; \textit{Joint predictives, quadratic-time}} \\
  TNPA & $1.898\pm0.005$ & $1.378\pm0.002$ & $0.701\pm0.001$ & $0.938\pm0.000$ & $(3.004\pm0.002){\times}10^{4}$ \\
  NDP  & $3.071\pm0.008$ & $1.753\pm0.002$ & $1.064\pm0.001$ & $0.590\pm0.000$ & $(9.154\pm0.015){\times}10^{3}$ \\
  FNP  & $2.395\pm0.008$ & $1.548\pm0.003$ & $0.839\pm0.001$ & $0.785\pm0.001$ & $(8.844\pm0.004){\times}10^{3}$ \\
\multicolumn{6}{@{}l}{\;\; \textit{Joint predictives, linear-time}} \\
  MFNP\textsuperscript{\dag} & $\mathbf{1.808\pm0.006}$ & $\mathbf{1.344\pm0.002}$ & $\mathbf{0.681\pm0.001}$ & $\mathbf{0.945\pm0.000}$ & $\mathbf{67.9\pm0.5}$ \\
\addlinespace

\midrule

\multicolumn{6}{l}{\textit{NOAA US} ($K{=}5{,}580$)} \\[2pt]
\multicolumn{6}{@{}l}{\;\; \textit{Joint predictives, quadratic-time}} \\
  TNPA & \textsc{t/o}     & \textsc{t/o}    & \textsc{t/o}    & \textsc{t/o} & \textsc{t/o} \\
  NDP  & $3.528\pm0.087$  & $1.878\pm0.023$ & $1.194\pm0.013$ & $0.322\pm0.005$ & $(7.075\pm0.020){\times}10^{5}$ \\
  FNP  & $2.872\pm0.064$& $1.695\pm0.019$ & $0.864\pm0.009$ & $0.786\pm0.001$ & $(8.831\pm0.002){\times}10^{5}$ \\
\multicolumn{6}{@{}l}{\;\; \textit{Joint predictives, linear-time}} \\
  MFNP\textsuperscript{\dag} & $\mathbf{2.554\pm0.010}$ & $\mathbf{1.598\pm0.003}$ & $\mathbf{0.774\pm0.001}$ & $\mathbf{0.955\pm0.001}$ & $\mathbf{(1.126\pm0.000){\times}10^{3}}$ \\
\bottomrule
\end{tabular}
\end{table}

\subsection{NOAA temperature interpolation}
\label{sec:noaa}

MFNP attains the lowest RMSE and CRPS on every tier and runs
$114$--$784{\times}$ faster than the strongest transformer
baseline (Table~\ref{tab:noaa-main}; factorised baselines and
per-tier Pareto frontiers in Appendix~\ref{app:full-noaa}).
Its $95\%$ predictive coverage is closest to nominal among
the evaluated joint-predictive models, reaching $0.940$ on Utah,
$0.945$ in California and $0.955$ across the continental US.
The respective contributions of distillation and architecture
to the speedup are examined in \S\ref{sec:speedup}.

Exact, sparse and nearest-neighbour GPs fit to each episode
serve as spatial references, with MSE $3.7$--$6.8{\times}$ that of
MFNP and Cov-95 of $0.92$--$0.94$ (Table~\ref{tab:noaa-gp}).
A representative prediction on the continental US is shown
in Figure~\ref{fig:noaa-qual-us}.

\subsection{Computational scaling}
\label{sec:scaling}
\label{sec:speedup}

\textbf{End-to-end sampling.}
MFNP reaches $K=10^{6}$ in under a second
on one NVIDIA L40S. We measure the time to generate one
joint predictive draw ($N_{\text{samples}}=1$, batch$=1$) on
2D Gaussian Mat\'ern-$\nicefrac{3}{2}$, using each method's
deployment configuration (Figure~\ref{fig:scaling},
Table~\ref{tab:scaling}).
FNP and NDP exhaust memory by $K=16{,}384$, and TNPA exhausts
memory by $K=65{,}536$. Even before reaching its memory limit,
TNPA takes ${\sim}117$\,s at $K=4{,}096$ and ${\sim}2$ hours
at $K=16{,}384$.

\textbf{Reflow versus architecture.}
To separate the effect of fewer sampling steps from the cost
of each forward pass, we compare the architectures at matched
1-NFE and parameter count
(Figure~\ref{fig:scaling-nfe1}, Appendix~\ref{app:distillation}).
MFNP, FNP, and NDP share a
${\sim}1.5$--$3.5$\,ms dispatch floor for $K\le1{,}024$, so the
order-of-magnitude speedup at small and moderate $K$ is attributable
to reflow distillation. At larger $K$, quadratic attention
becomes more expensive and eventually exceeds the memory budget,
while MFNP retains linear-time computation. Both effects contribute
to the $628$--$784{\times}$ speedup over NDP and FNP on NOAA US
(Table~\ref{tab:noaa-main}).
At the US tier, NDP and FNP fit the per-GPU memory budget only via
sample-axis chunking, which preserves their predictive distribution
while retaining the quadratic per-step cost in the point count.

\subsection{Ablations}
\label{sec:ablation-results}

Each row of Table~\ref{tab:ablation} changes one model
component (Appendix~\ref{app:ablations}).

\textbf{Context compression.}
Compressing the context into a single state gives $10$--$17\%$
higher MSE than the full model on synthetic tasks and $58$--$65\%$
higher MSE on NOAA. These results support decoding queries
from states at their own scan positions rather than from one
terminal context state.

\textbf{Spatial routing.}
Random ordering raises MSE by $1.2$--$1.7{\times}$ on synthetic tasks
and $2.0$--$5.0{\times}$ on NOAA (Table~\ref{tab:ablation}).
Reducing the $4$-direction routing to one direction inflates MSE by
$1.6$--$4.2{\times}$; using two directions increases MSE
by $3$--$9\%$ on the headline NOAA tiers while using
$31$--$74\%$ of the runtime, and matches or improves on the full model's synthetic MSE.

Morton and lexicographic curves give $4\%$ and $9\%$ lower MSE than the
Hilbert curve on 2D synthetic data, and lexicographic ordering raises CRPS by
$16\%$
(Table~\ref{tab:ordering}). The boundary diagnostic gives
comparable boundary-to-interior MSE ratios for MFNP and FNP, with
$1.40$ versus $1.54$ on Utah and $1.43$ versus $1.47$ on California
(Appendix~\ref{app:boundary}).

\textbf{Sampling.} The deployment configuration uses one-step Euler sampling
from the reflow-distilled student. The teacher--student and solver
comparisons are reported in Table~\ref{tab:distillation}
(Appendix~\ref{app:distillation}).
On the student, multi-step solvers do not reduce MSE on any task.
On 2D Mat\'ern-$\nicefrac{3}{2}$, $15$-step Euler gives an MSE of $0.0166$
and a CRPS of $0.0537$, against $0.0167$ and $0.0537$ for one-step Euler,
and $15$- and $30$-step Euler take $14$ and $28$ times as long.
Heun-1 inflates NOAA MSE by $1.58$--$1.86{\times}$
over Euler-1 and is not used.

\section{Conclusion}
\label{sec:conclusion}

MambaFlowNP combines joint prediction on irregular spatial
point sets with $O(K)$ computation per forward pass. Reflow
distillation enables one-pass sampling. Across the three NOAA
tiers, MFNP attains the lowest RMSE and CRPS among the
evaluated joint-predictive models, with near-nominal coverage and
$114$--$784{\times}$ lower latency than the strongest transformer
baseline. On synthetic Gaussian processes, it matches TNPA's MSE
within one SE on 1D Mat\'ern-$\nicefrac{3}{2}$ and runs $25$--$42{\times}$
faster than TNPA and FNP on Mat\'ern-$\nicefrac{3}{2}$.
A single forward pass at $10^{6}$ points completes in under a
second on one NVIDIA L40S GPU, demonstrating the
computational feasibility of joint prediction over large,
irregular spatial point sets.

\paragraph{Limitations.}
MFNP targets irregular point sets. On regularly gridded domains,
convolutional NPs and FFT-based GP solvers exploit grid structure
directly and are typically more efficient. At small $K$
($\lesssim 10^{3}$) MFNP's per-call dispatch floor leaves it slower
than the simplest factorised NPs (Tables~\ref{tab:synth-full} and~\ref{tab:noaa-full}). Its scaling advantage over
quadratic attention emerges at larger point counts. We evaluate one- and
two-dimensional inputs only; space--time inputs need one extra coordinate in the
Hilbert ordering, such as a three-dimensional curve over $(x, y, t)$, and we leave
their evaluation to future work.

Reflow distillation adds a coupled-pair generation pass to the
training pipeline that single-stage baselines do not incur.
The reported inference speedups combine the effects of
distillation and the architecture.

Our real-data evaluation is limited to one dataset and one
variable, NOAA daily maximum temperature. The split is temporal
within a fixed station network, and transfer to held-out regions
remains untested. Evaluation against scalable spatial baselines
is incomplete, as structured kernel interpolation, local kriging,
graph-based methods and AR-ConvCNP~\citep{bruinsma2023arcnp}
remain to be tested. Predictive quality at $10^{6}$ points is
also unestablished.

\paragraph{Code and reproducibility.}
MambaFlowNP code, trained weights, and training and evaluation
scripts for the three NOAA tiers are available at
\url{https://github.com/CosmoNaught/MambaFlowNP}.

\begin{ack}
CS was supported by the Wellcome Trust (226072/Z/22/Z). GC's
research was funded in whole, or in part, by the Wellcome Trust
(220900/Z/20/Z). TJH and OJW
acknowledge support from the Academy of Medical Sciences, the
Wellcome Trust, the Government Department of Science, Innovation and
Technology, the British Heart Foundation and Diabetes UK (Springboard
Award SBF0010\textbackslash{}1221). OJW is also supported by an
Imperial College Research Fellowship sponsored by Schmidt Sciences.
ES acknowledges support in part from the AI2050 program at Schmidt
Sciences (Grant G-22-64476). All authors acknowledge funding from the
Medical Research Council (MRC) Centre for Global Infectious Disease
Analysis (reference MR/X020258/1), funded by the United Kingdom MRC.
We thank Richard E. Turner and Cristiana-Diana Diaconu for generously
sharing their time and for insightful discussions on neural processes
and state-space models during the development of this work. We thank
the Research Software for Infectious Disease Epidemiology
(RESIDE) team at Imperial College London for access to their
computing cluster. The GPU compute was funded by MRC Mid-Range
Equipment Funding (UKRI2681). For the purpose of open access,
the authors have applied a CC BY public copyright licence to any
Author Accepted Manuscript version arising from this submission.
\end{ack}

\clearpage
\bibliographystyle{plainnat}
\bibliography{refs}


\appendix

\section{Hyperparameters}
\label{app:hparams}

All models train with AdamW (weight decay $10^{-4}$, the optax default; gradient clip $1.0$), a warmup-cosine learning-rate schedule (linear warmup over $\min(100, T{-}1)$ steps, cosine decay to $10^{-5}$), and EMA weights ($\beta{=}0.9995$).
Per-model hyperparameters come from a two-phase Bayesian-optimisation sweep, with a $16$-trial coarse pass plus a $16$-trial narrowed pass around the coarse optimum, giving $32$ trials per (model, task) cell, each optimising validation MSE.
On the synthetic tasks, MFNP is tuned by one Bayesian-optimisation sweep of $41$ to $50$ trials per task over hidden width, state size, depth, RFF frequencies and bandwidth, learning rate and batch size, each trial trained for $20$k flow-matching and $5$k reflow steps and scored by validation MSE on $64$ fixed episodes drawn from the task prior, and the best configuration is retrained with the full budget.

\paragraph{Latency methodology (\S\ref{sec:protocol}).} Baselines
retain their authors' attention implementations; the additive
context--query mask routes PyTorch SDPA to the memory-efficient
kernel rather than FlashAttention, so reported OOM thresholds
reflect $O(K^2)$-memory attention. TNPA's autoregressive decoder runs
$M$ sequential self-attention passes per draw and is serial within
a draw, so sample chunking cannot reduce its cost; on NOAA US the
projected per-seed runtime exceeds the per-cell budget by an order
of magnitude.

\paragraph{NOAA preprocessing (\S\ref{sec:tasks}).} The Utah tier
spans $37$--$42^{\circ}$N, $109$--$114^{\circ}$W; California spans
$32$--$42^{\circ}$N, $114$--$124.5^{\circ}$W. Stations are kept if
they report on at least $80\%$ of train-window days, with top-coverage
fallback to the per-tier cap. Station coordinates are
equirectangular-projected to kilometres and axis-wise min-max scaled
to $[0,1]^2$; observations pass through a Gaussian quantile transform
fit on the train window, with metrics reported on the denormalised
\textdegree C scale.

\begin{table}[!htbp]
\centering
\caption{Best per-cell hyperparameters on the 1D synthetic Gaussian tasks ($K{=}128$). \textbf{Architecture.} $d_{\text{model}}$ hidden width; $d_{\text{state/lat}}$ Mamba state / NP latent width; $L$ blocks; $h{\,/\,}F_\phi$ transformer heads or MFNP RFF frequencies; $d_{\text{ff}}$ transformer FFN width. \textbf{Training.} AdamW lr $\eta$, batch $|B|$, optimiser steps (CFM teacher\,$+$\,reflow for MFNP). \textbf{Inference.} $T_{\text{NFE}}$ integration / diffusion steps; $g_{\text{scale}}$ NDP guidance scale (jointly tuned with architecture in BO on synthetic tasks, and set by a post-hoc grid sweep on a fixed checkpoint on NOAA, selected to minimise validation MSE). Dashes mark N/A. \textsuperscript{\dag} our method.}
\label{tab:hparams-1d}
\scriptsize
\setlength{\tabcolsep}{3pt}
\resizebox{\textwidth}{!}{%
\begin{tabular}{@{} l c c c c c c c c c c @{}}
\toprule
 & \multicolumn{5}{c}{\textbf{Architecture}} & \multicolumn{3}{c}{\textbf{Training}} & \multicolumn{2}{c}{\textbf{Inference}} \\
\cmidrule(lr){2-6} \cmidrule(lr){7-9} \cmidrule(lr){10-11}
\textbf{Model} & $d_{\text{model}}$ & $d_{\text{state/lat}}$ & $L$ & $h{\,/\,}F_\phi$ & $d_{\text{ff}}$ & $\eta$ & $|B|$ & \textbf{steps} & $T_{\text{NFE}}$ & $g_{\text{scale}}$ \\
\midrule

\multicolumn{11}{l}{\textit{1D Gaussian -- RBF} ($K{=}128$)} \\[2pt]
\multicolumn{11}{@{}l}{\;\; \textit{Factorised predictives}} \\
  CNP  & $256$ & --    & -- & --  & --    & $2.6{\times}10^{-3}$ & $64$ & $100$k             & --    & --      \\
  NP   & $256$ & $256$ & -- & --  & --    & $2.2{\times}10^{-3}$ & $64$ & $100$k             & --    & --      \\
  ANP  & $256$ & $128$ & -- & --  & --    & $3.0{\times}10^{-4}$ & $64$ & $100$k             & --    & --      \\
  CANP & $ 64$ & --    & -- & --  & --    & $1.9{\times}10^{-3}$ & $64$ & $100$k             & --    & --      \\
\multicolumn{11}{@{}l}{\;\; \textit{Joint predictives, quadratic-time}} \\
  TNPA & $128$ & --    & $6$ & $4$ & $128$ & $7.2{\times}10^{-4}$ & $32$ & $100$k             & --    & --      \\
  NDP  & $128$ & --    & $6$ & $8$ & $128$ & $8.4{\times}10^{-4}$ & $16$ & $100$k             & $100$ & $400$   \\
  FNP  & $256$ & --    & $8$ & $4$ & $256$ & $2.7{\times}10^{-4}$ & $64$ & $100$k             & $100$ & --      \\
\multicolumn{11}{@{}l}{\;\; \textit{Joint predictives, linear-time}} \\
  MFNP\textsuperscript{\dag} & $384$ & $32$ & $3$ & $128$ & -- & $8.4{\times}10^{-4}$ & $128$ & $80$k\,$+$\,$20$k & $1$ & -- \\
\addlinespace

\midrule

\multicolumn{11}{l}{\textit{1D Gaussian -- Mat\'ern-$\nicefrac{3}{2}$} ($K{=}128$)} \\[2pt]
\multicolumn{11}{@{}l}{\;\; \textit{Factorised predictives}} \\
  CNP  & $256$ & --    & -- & --  & --    & $3.1{\times}10^{-3}$ & $64$ & $100$k             & --    & --      \\
  NP   & $128$ & $256$ & -- & --  & --    & $2.9{\times}10^{-3}$ & $64$ & $100$k             & --    & --      \\
  ANP  & $ 64$ & $128$ & -- & --  & --    & $1.2{\times}10^{-3}$ & $32$ & $100$k             & --    & --      \\
  CANP & $256$ & --    & -- & --  & --    & $7.1{\times}10^{-4}$ & $16$ & $100$k             & --    & --      \\
\multicolumn{11}{@{}l}{\;\; \textit{Joint predictives, quadratic-time}} \\
  TNPA & $ 64$ & --    & $6$ & $8$ & $512$ & $1.6{\times}10^{-4}$ & $32$ & $100$k             & --    & --      \\
  NDP  & $256$ & --    & $8$ & $4$ & $128$ & $3.4{\times}10^{-4}$ & $64$ & $100$k             & $100$ & $900$   \\
  FNP  & $ 64$ & --    & $8$ & $8$ & $256$ & $5.3{\times}10^{-4}$ & $64$ & $100$k             & $100$ & --      \\
\multicolumn{11}{@{}l}{\;\; \textit{Joint predictives, linear-time}} \\
  MFNP\textsuperscript{\dag} & $128$ & $64$ & $8$ & $192$ & -- & $9.9{\times}10^{-4}$ & $128$ & $80$k\,$+$\,$20$k & $1$ & -- \\
\addlinespace

\midrule

\multicolumn{11}{l}{\textit{1D Gaussian -- Mat\'ern-$\nicefrac{5}{2}$} ($K{=}128$)} \\[2pt]
\multicolumn{11}{@{}l}{\;\; \textit{Factorised predictives}} \\
  CNP  & $256$ & --    & -- & --  & --    & $1.1{\times}10^{-3}$ & $64$ & $100$k             & --    & --          \\
  NP   & $128$ & $128$ & -- & --  & --    & $2.5{\times}10^{-3}$ & $64$ & $100$k             & --    & --          \\
  ANP  & $128$ & $256$ & -- & --  & --    & $1.4{\times}10^{-3}$ & $64$ & $100$k             & --    & --          \\
  CANP & $ 64$ & --    & -- & --  & --    & $2.1{\times}10^{-3}$ & $64$ & $100$k             & --    & --          \\
\multicolumn{11}{@{}l}{\;\; \textit{Joint predictives, quadratic-time}} \\
  TNPA & $256$ & --    & $8$ & $8$ & $512$ & $1.4{\times}10^{-4}$ & $16$ & $100$k             & --    & --          \\
  NDP  & $256$ & --    & $8$ & $8$ & $512$ & $1.1{\times}10^{-4}$ & $32$ & $100$k             & $100$ & $2{,}200$   \\
  FNP  & $128$ & --    & $4$ & $4$ & $256$ & $2.6{\times}10^{-4}$ & $16$ & $100$k             & $100$ & --          \\
\multicolumn{11}{@{}l}{\;\; \textit{Joint predictives, linear-time}} \\
  MFNP\textsuperscript{\dag} & $128$ & $16$ & $3$ & $192$ & -- & $1.6{\times}10^{-3}$ & $128$ & $80$k\,$+$\,$20$k & $1$ & -- \\
\bottomrule
\end{tabular}%
}
\end{table}

\begin{table}[!htbp]
\centering
\caption{Best per-cell hyperparameters on the 2D synthetic Gaussian Mat\'ern-$\nicefrac{3}{2}$ task ($K{=}256$). Column conventions as in Table~\ref{tab:hparams-1d}; factorised baselines are not reported on this task. \textsuperscript{\dag} our method.}
\label{tab:hparams-2d}
\scriptsize
\setlength{\tabcolsep}{3pt}
\resizebox{\textwidth}{!}{%
\begin{tabular}{@{} l c c c c c c c c c c @{}}
\toprule
 & \multicolumn{5}{c}{\textbf{Architecture}} & \multicolumn{3}{c}{\textbf{Training}} & \multicolumn{2}{c}{\textbf{Inference}} \\
\cmidrule(lr){2-6} \cmidrule(lr){7-9} \cmidrule(lr){10-11}
\textbf{Model} & $d_{\text{model}}$ & $d_{\text{state/lat}}$ & $L$ & $h{\,/\,}F_\phi$ & $d_{\text{ff}}$ & $\eta$ & $|B|$ & \textbf{steps} & $T_{\text{NFE}}$ & $g_{\text{scale}}$ \\
\midrule
\multicolumn{11}{l}{\textit{2D Gaussian -- Mat\'ern-$\nicefrac{3}{2}$} ($K{=}256$)} \\[2pt]
\multicolumn{11}{@{}l}{\;\; \textit{Joint predictives, quadratic-time}} \\
  TNPA & $128$ & --    & $6$ & $4$ & $512$ & $4.6{\times}10^{-4}$ & $64$ & $100$k             & --    & --          \\
  NDP  & $256$ & --    & $4$ & $4$ & $256$ & $4.3{\times}10^{-4}$ & $32$ & $100$k             & $100$ & $6{,}500$   \\
  FNP  & $ 64$ & --    & $6$ & $8$ & $256$ & $7.4{\times}10^{-4}$ & $32$ & $100$k             & $100$ & --          \\
\multicolumn{11}{@{}l}{\;\; \textit{Joint predictives, linear-time}} \\
  MFNP\textsuperscript{\dag} & $256$ & $128$ & $6$ & $384$ & -- & $1.1{\times}10^{-3}$ & $64$ & $80$k\,$+$\,$20$k & $1$ & -- \\
\bottomrule
\end{tabular}%
}
\end{table}

\begin{table}[!htbp]
\centering
\caption{Best per-cell hyperparameters on the three NOAA tiers (Utah $K{=}222$; California $K{=}486$; continental US $K{=}5{,}580$). Column conventions as in Table~\ref{tab:hparams-1d}. TNPA exceeds the per-cell compute budget on the US tier (\S\ref{sec:protocol}) and is omitted there. MFNP uses $4$ Mamba-2 heads and a scan chunk length of $64$. \textsuperscript{\dag} our method.}
\label{tab:hparams-noaa}
\scriptsize
\setlength{\tabcolsep}{3pt}
\resizebox{\textwidth}{!}{%
\begin{tabular}{@{} l c c c c c c c c c c @{}}
\toprule
 & \multicolumn{5}{c}{\textbf{Architecture}} & \multicolumn{3}{c}{\textbf{Training}} & \multicolumn{2}{c}{\textbf{Inference}} \\
\cmidrule(lr){2-6} \cmidrule(lr){7-9} \cmidrule(lr){10-11}
\textbf{Model} & $d_{\text{model}}$ & $d_{\text{state/lat}}$ & $L$ & $h{\,/\,}F_\phi$ & $d_{\text{ff}}$ & $\eta$ & $|B|$ & \textbf{steps} & $T_{\text{NFE}}$ & $g_{\text{scale}}$ \\
\midrule

\multicolumn{11}{l}{\textit{NOAA Utah} ($K{=}222$)} \\[2pt]
\multicolumn{11}{@{}l}{\;\; \textit{Factorised predictives}} \\
  CNP  & $256$ & --    & -- & --  & --    & $1.9{\times}10^{-3}$ & $ 8$ & $100$k             & --    & --         \\
  NP   & $256$ & $ 64$ & -- & --  & --    & $3.7{\times}10^{-3}$ & $ 8$ & $100$k             & --    & --         \\
  ANP  & $128$ & $ 64$ & -- & --  & --    & $1.2{\times}10^{-3}$ & $ 4$ & $100$k             & --    & --         \\
  CANP & $256$ & --    & -- & --  & --    & $9.4{\times}10^{-4}$ & $ 4$ & $100$k             & --    & --         \\
\multicolumn{11}{@{}l}{\;\; \textit{Joint predictives, quadratic-time}} \\
  TNPA & $128$ & --    & $8$ & $8$ & $512$ & $2.9{\times}10^{-4}$ & $ 4$ & $100$k             & --    & --         \\
  NDP  & $256$ & --    & $3$ & $4$ & $512$ & $2.0{\times}10^{-4}$ & $ 8$ & $100$k             & $100$ & $25{,}000$ \\
  FNP  & $256$ & --    & $4$ & $8$ & $768$ & $1.3{\times}10^{-4}$ & $ 4$ & $100$k             & $100$ & --         \\
\multicolumn{11}{@{}l}{\;\; \textit{Joint predictives, linear-time}} \\
  MFNP\textsuperscript{\dag} & $128$ & $16$ & $6$ & $256$ & -- & $1.0{\times}10^{-3}$ & $16$ & $80$k\,$+$\,$20$k & $1$ & -- \\
\addlinespace

\midrule

\multicolumn{11}{l}{\textit{NOAA California} ($K{=}486$)} \\[2pt]
\multicolumn{11}{@{}l}{\;\; \textit{Factorised predictives}} \\
  CNP  & $256$ & --    & --  & --  & --    & $1.4{\times}10^{-3}$ & $ 8$ & $100$k             & --    & --         \\
  NP   & $256$ & $ 64$ & --  & --  & --    & $2.0{\times}10^{-3}$ & $ 8$ & $100$k             & --    & --         \\
  ANP  & $256$ & $128$ & --  & --  & --    & $1.2{\times}10^{-3}$ & $ 4$ & $100$k             & --    & --         \\
  CANP & $256$ & --    & --  & --  & --    & $9.3{\times}10^{-4}$ & $ 4$ & $100$k             & --    & --         \\
\multicolumn{11}{@{}l}{\;\; \textit{Joint predictives, quadratic-time}} \\
  TNPA & $128$ & --    & $10$ & $4$ & $512$ & $1.8{\times}10^{-4}$ & $ 4$ & $100$k             & --    & --         \\
  NDP  & $256$ & --    & $4$  & $8$ & $256$ & $1.9{\times}10^{-4}$ & $ 8$ & $100$k             & $100$ & $25{,}000$ \\
  FNP  & $128$ & --    & $4$  & $8$ & $768$ & $2.1{\times}10^{-4}$ & $ 2$ & $100$k             & $100$ & --         \\
\multicolumn{11}{@{}l}{\;\; \textit{Joint predictives, linear-time}} \\
  MFNP\textsuperscript{\dag} & $256$ & $16$ & $6$ & $192$ & -- & $1.5{\times}10^{-3}$ & $16$ & $80$k\,$+$\,$20$k & $1$ & -- \\
\addlinespace

\midrule

\multicolumn{11}{l}{\textit{NOAA US} ($K{=}5{,}580$)} \\[2pt]
\multicolumn{11}{@{}l}{\;\; \textit{Factorised predictives}} \\
  CNP  & $768$ & --    & -- & --  & --    & $1.3{\times}10^{-3}$ & $32$ & $100$k             & --    & --         \\
  NP   & $512$ & $384$ & -- & --  & --    & $9.4{\times}10^{-4}$ & $32$ & $100$k             & --    & --         \\
  ANP  & $256$ & $256$ & -- & --  & --    & $1.4{\times}10^{-3}$ & $ 2$ & $100$k             & --    & --         \\
  CANP & $256$ & --    & -- & --  & --    & $5.3{\times}10^{-4}$ & $ 4$ & $100$k             & --    & --         \\
\multicolumn{11}{@{}l}{\;\; \textit{Joint predictives, quadratic-time}} \\
  NDP  & $384$ & --    & $3$ & $8$ & $512$ & $1.4{\times}10^{-4}$ & $16$ & $100$k             & $100$ & $15{,}000$ \\
  FNP  & $256$ & --    & $4$ & $4$ & $512$ & $3.5{\times}10^{-4}$ & $ 4$ & $100$k             & $100$ & --         \\
\multicolumn{11}{@{}l}{\;\; \textit{Joint predictives, linear-time}} \\
  MFNP\textsuperscript{\dag} & $256$ & $16$ & $6$ & $384$ & -- & $1.0{\times}10^{-3}$ & $ 8$ & $80$k\,$+$\,$20$k & $1$ & -- \\
\bottomrule
\end{tabular}%
}
\end{table}

\section{Detailed algorithms}
\label{app:detailed-algs}

We first specify the feature maps, routing settings and scan
implementation, then give the forward pass, CFM training step and
sampling procedure. The coordinate embedding $\phi$ and the
random-function feature map $\boldsymbol{\psi}$ serve different
roles, as $\phi$ encodes each input location, while
$\boldsymbol{\psi}$ defines the initial query values.

\paragraph{Embedding (\S\ref{sec:method}).}
The frozen Random Fourier Features map~\citep{rahimi2007random}
referenced as $\phi$ in Algorithm~\ref{alg:MFNP_detailed} is
\begin{equation}
\begin{aligned}
\phi(\mathbf{x})
    &= \bigl[\cos(2\pi B\mathbf{x}),\;
             \sin(2\pi B\mathbf{x})\bigr],\\
B &\in \mathbb{R}^{F_\phi \times 2},\qquad
B_{jk} \stackrel{\text{iid}}{\sim}
    \mathcal{N}(0,\sigma_\phi^2),\\
\sigma_\phi &= 0.25/d_{\min},
\end{aligned}
\end{equation}
with $d_{\min}=0.5/\sqrt{K_{\max}}$ and $F_\phi$ the number
of coordinate-embedding frequencies. This is separate from the
source feature dimension $F=256$ below.

\paragraph{Random-function source distribution
(\S\ref{sec:method}).}
The source of each draw is
$Y_{0,i}
=\boldsymbol{\psi}(\mathbf{x}_{Q,i})^{\top}\mathbf{z}$
with $\mathbf{z}\sim\mathcal{N}(0,I_F)$, $F=256$, and
\begin{equation}
\begin{aligned}
\boldsymbol{\psi}(\mathbf{x})
    &= \sqrt{\tfrac{2}{F}}\,
       \bigl[\cos(\boldsymbol{\nu}_1^{\!\top}\mathbf{x}+c_1),
       \ldots,
       \cos(\boldsymbol{\nu}_F^{\!\top}\mathbf{x}+c_F)
       \bigr]^{\!\top},\\
\boldsymbol{\nu}_j
    &\stackrel{\text{iid}}{\sim}
      \mathcal{N}(0,K_{\max}I_2),\qquad
c_j \stackrel{\text{iid}}{\sim}\mathcal{U}(0,2\pi),
\end{aligned}
\end{equation}
fixed at initialisation. $\mathbf{Y}_0$ is then an approximate
draw from a unit-variance Gaussian process with RBF kernel and
lengthscale $1/\sqrt{K_{\max}}$, and the same $\mathbf{z}$
conditions every block through adaLN.

\paragraph{Routing settings.}
Let $\mathcal{H}: [0,1]^2 \to \mathbb{N}$ be the index function of
a $16$th-order curve, which quantises each point to one of $2^{32}$
lattice cells. Let $\mathbf{s}_i \in \mathbb{R}^2$ be the spatial
coordinate of the $i$th point in the concatenation
$[\mathbf{X}_C \,\|\, \mathbf{X}_Q]$.
MFNP applies the curve to the shifted coordinates
$\tfrac{1}{2}\mathbf{s}_i+\mathbf{o}$, with a fresh offset
$\mathbf{o}\sim\mathcal{U}[0,\tfrac{1}{2})^2$ for every training
episode. At inference, the $N_{\text{samples}}$ draws are split evenly
across $8$ offsets and pooled. Distances $\delta_{d,i}$ use the
unshifted coordinates.
Each direction $d$ caches the distance $\delta_{d,i}$ from
point $\pi_d(i)$ to the nearest preceding context point in that
order, scaled by $\sqrt{N}$. The sort and distance cache depend
only on $(\mathbf{X}_C,\mathbf{X}_Q)$ and the routing offset
$\mathbf{o}$.

For one-dimensional inputs, points are sorted by their scalar
coordinate and scanned in ascending and descending order, so the
ensemble has two distinct directions. The four scan slots repeat
these two orderings, as described in Appendix~\ref{app:ablations}.
Algorithm~\ref{alg:MFNP_detailed} shows the two-dimensional routing.

\paragraph{Scan implementation.}
Context tokens write to the recurrent state. Query tokens read
the state at their own positions but do not write to it. The
distance vector $\delta_d$ is concatenated to the input of each
block's $\Delta t$ gate, whose softplus output is clipped to
$[10^{-3},10]$. The gated scan output is RMS-normalised before the
output projection.
We use the JAX/XLA native implementation
of~\citet{santoni2026compiler}.

\begin{algorithm}[!htb]
\caption{MambaFlowNP forward pass (detailed).}
\label{alg:MFNP_detailed}
\small
\begin{algorithmic}[1]
\Require Context $\mathbf{X}_C\in\mathbb{R}^{N\times D}$,
    $\mathbf{Y}_C\in\mathbb{R}^N$.
\Require Query coordinates
    $\mathbf{X}_Q\in\mathbb{R}^{M\times D}$,
    noisy query slice $\mathbf{Y}_t\in\mathbb{R}^M$,
    flow time $t\in[0,1]$, draw latent
    $\mathbf{z}\in\mathbb{R}^F$, routing offset
    $\mathbf{o}\in[0,\tfrac{1}{2})^2$.
\Require Frozen RFF projection $\phi$; optional routing cache.

\State Let $K=N+M$. Concatenate coordinates
    $\mathbf{S}=[\mathbf{X}_C\,\|\,\mathbf{X}_Q]
    \in\mathbb{R}^{K\times D}$.

\State \textbf{Phase 1: Tokenisation \& Conditioning}
\State $\mathbf{e}_t \gets
    \text{MLP}_{\text{time}}(\text{SinusoidalEmbedding}(t))$
    ${}+\text{MLP}_{\text{lat}}(\mathbf{z})$
\For{$i=1$ \textbf{to} $K$}
    \If{$i\le N$}
        \State $\mathbf{E}_i \gets
            W_{\text{ctx}}\,[\phi(\mathbf{s}_i),Y_{C,i}]$
    \Else
        \State $\mathbf{E}_i \gets
            W_{\text{qry}}\,[\phi(\mathbf{s}_i),Y_{t,i-N}]$
    \EndIf
\EndFor

\State \textbf{Phase 2: Multi-directional routing}
    (pre-computed, cached)
\Statex Reuse the cached quantities if supplied;
    otherwise compute the following.
\State $\tilde{\mathbf{S}}\gets
    \tfrac{1}{2}\mathbf{S}+\mathbf{o}$
    \Comment{jittered scan grid}
\State $\pi_1\gets\text{argsort}
    \bigl(\mathcal{H}(\tilde{\mathbf{S}}_{:,1},
    \tilde{\mathbf{S}}_{:,2})\bigr)$
\State $\pi_2\gets\text{argsort}
    \bigl(\mathcal{H}(\tilde{\mathbf{S}}_{:,2},
    \tilde{\mathbf{S}}_{:,1})\bigr)$
\State Define ensemble
    $\Pi=\{\pi_1,\pi_1^{\mathrm{rev}},
    \pi_2,\pi_2^{\mathrm{rev}}\}$.
\State Compute the inverse permutation
    $\pi_d^{-1}\gets\text{argsort}(\pi_d)$ for each $d\in\Pi$.
\State $\delta_{d,i}\gets
    \sqrt{N}\,\lVert
    \mathbf{s}_{\pi_d(i)}-\mathbf{s}_{\pi_d(j)}
    \rVert$ for each $d,i$, with $j<i$ the nearest preceding
    context position in $\pi_d$
    ($\delta_{d,i}=0$ if none).

\State \textbf{Phase 3: Mamba-2 with distance-aware adaLN-Zero}
\State Apply permutations:
    $\mathbf{E}_d\gets\mathbf{E}[\pi_d]$ for each $d$.
\For{$\ell=1$ \textbf{to} $L$}
    \State $(\gamma_\ell,\beta_\ell,\alpha_\ell)
        \gets W^{(\ell)}_{\text{adaln}}\,
        \mathrm{SiLU}(\mathbf{e}_t)$
        \Comment{zero-init scale, shift, gate}
    \State $\mathbf{E}_d\gets
        \mathbf{E}_d+\alpha_\ell\odot
        \text{Mamba2}\!\bigl(
        \text{LN}(\mathbf{E}_d)\odot(1+\gamma_\ell)
        +\beta_\ell,\;\delta_d;\;\theta\bigr)$
        for each $d$.
\EndFor

\State \textbf{Phase 4: Fusion \& Readout}
\State $\mathbf{H}_d\gets
    \mathbf{E}_d[\pi_d^{-1}]$ for each $d$.
    \Comment{inverse permutation}
\State $\mathbf{H}\gets W_{\text{fuse}}
    [\mathbf{H}_{\pi_1}\,\|\,
    \mathbf{H}_{\pi_1^{\mathrm{rev}}}\,\|\,
    \mathbf{H}_{\pi_2}\,\|\,
    \mathbf{H}_{\pi_2^{\mathrm{rev}}}]
    +b_{\text{fuse}}$
\State $\hat{\mathbf{v}}\gets
    W_{\text{out}}\bigl(
    \text{adaLN}(\text{LN}(\mathbf{H}[N+1:K]),
    \mathbf{e}_t)\bigr)+b_{\text{out}}$
    \Comment{zero-init output}
\State \Return $\hat{\mathbf{v}}$
\end{algorithmic}
\end{algorithm}

\begin{algorithm}[!htb]
\caption{MambaFlowNP CFM training step.}
\label{alg:MFNP_train_detailed}
\small
\begin{algorithmic}[1]
\Require Episode sampler returning
    $(\mathbf{X}_C,\mathbf{Y}_C,\mathbf{X}_Q,\mathbf{Y}_Q)$;
    parameters $\theta$.
\State Sample episode and $t\sim\mathcal{U}(0,1)$.
\State Draw $\mathbf{z}\sim\mathcal{N}(0,I_F)$,
    $\mathbf{o}\sim\mathcal{U}[0,\tfrac{1}{2})^2$, and set
    $Y_{0,i}\gets
    \boldsymbol{\psi}(\mathbf{x}_{Q,i})^\top\mathbf{z}$.
\State $\mathbf{Y}_t\gets
    (1-t)\mathbf{Y}_0+t\mathbf{Y}_Q$;
    \quad $\mathbf{u}^{\star}\gets\mathbf{Y}_Q-\mathbf{Y}_0$.
\State $\hat{\mathbf{v}}\gets
    \text{MFNP}_\theta(
    \mathbf{X}_C,\mathbf{Y}_C,\mathbf{X}_Q,\mathbf{Y}_t,t
    ,\mathbf{z},\mathbf{o})$
    via Algorithm~\ref{alg:MFNP_detailed}.
\State $\mathcal{L}\gets
    \lVert\hat{\mathbf{v}}-\mathbf{u}^{\star}\rVert^2$.
    Update $\theta$ by gradient step on $\mathcal{L}$.
\end{algorithmic}
\end{algorithm}

\paragraph{Sampling and clipping.}
Algorithm~\ref{alg:MFNP_inference_detailed} generates one
joint draw. The latent $\mathbf{z}$ and routing offset
$\mathbf{o}$ remain fixed throughout its ODE integration.
A one-step Euler call is the deployment sampler.
The clipping bound $y_{\max}$ is the support limit of the
per-task quantile-transformed targets ($y_{\max}=6$ on synthetic
tasks and on NOAA after the train-window quantile transform of
\S\ref{sec:tasks}); clipping is inactive on every reported run
and is retained for numerical safety only.

\begin{algorithm}[!htb]
\caption{Posterior sampling (detailed; Heun / Euler).}
\label{alg:MFNP_inference_detailed}
\small
\begin{algorithmic}[1]
\Require Context $\mathbf{X}_C,\mathbf{Y}_C$,
    queries $\mathbf{X}_Q$, trained MFNP
    (Algorithm~\ref{alg:MFNP_detailed}),
    steps $N_{\text{steps}}$, solver $\in\{\text{Euler},\text{Heun}\}$.
\State Draw the latent
    $\mathbf{z}\sim\mathcal{N}(0,I_F)$ and the routing offset
    $\mathbf{o}$ of this draw (one of $8$ offsets shared by
    $N_{\text{samples}}/8$ draws).
\State Pre-compute routing cache from
    $(\mathbf{X}_C,\mathbf{X}_Q,\mathbf{o})$:
    permutations $\{\pi_d:d\in\Pi\}$ with
    $\Pi=\{\pi_1,\pi_1^{\mathrm{rev}},
    \pi_2,\pi_2^{\mathrm{rev}}\}$,
    inverse permutations $\{\pi_d^{-1}\}$,
    and distances $\{\delta_d\}$.
\State Set $Y_{0,i}\gets
    \boldsymbol{\psi}(\mathbf{x}_{Q,i})^\top\mathbf{z}$,
    the random-function source of \S\ref{sec:method}.
\State $\Delta t\gets 1/N_{\text{steps}}$.
\For{$k=0$ \textbf{to} $N_{\text{steps}}-1$}
    \State $t_1\gets k\Delta t$;
        \quad $t_2\gets(k+1)\Delta t$.
    \State $\mathbf{v}_1\gets
        \text{MFNP}(\mathbf{X}_C,\mathbf{Y}_C,\mathbf{X}_Q,
        \mathbf{Y}_{t_1},t_1,\mathbf{z},\mathbf{o};
        \text{cache})$
    \If{Heun}
        \State $\tilde{\mathbf{Y}}\gets
            \text{clip}_{[-y_{\max},y_{\max}]}
            \!\bigl(\mathbf{Y}_{t_1}
            +\Delta t\cdot\mathbf{v}_1\bigr)$
        \State $\mathbf{v}_2\gets
            \text{MFNP}(\mathbf{X}_C,\mathbf{Y}_C,\mathbf{X}_Q,
            \tilde{\mathbf{Y}},t_2,\mathbf{z},\mathbf{o};
            \text{cache})$
        \State $\mathbf{Y}_{t_2}\gets
            \text{clip}_{[-y_{\max},y_{\max}]}
            \!\bigl(\mathbf{Y}_{t_1}
            +\tfrac{\Delta t}{2}
            (\mathbf{v}_1+\mathbf{v}_2)\bigr)$
    \Else
        \State $\mathbf{Y}_{t_2}\gets
            \mathbf{Y}_{t_1}+\Delta t\cdot\mathbf{v}_1$
    \EndIf
\EndFor
\State \Return $\mathbf{Y}_1$
    \Comment{$\hat{\mathbf{Y}}_1$, a posterior sample}
\end{algorithmic}
\end{algorithm}

\section{Sort scaling}
\label{app:sort-scaling}

Table~\ref{tab:sort-scaling} reports wall-clock for the Hilbert-key
\texttt{argsort} call against $K$, alongside a same-shape no-op
identity baseline isolating fixed per-call cost (kernel launch,
dispatch) from $K$-dependent work, and the MFNP 1-NFE Euler
forward-pass wall-clock at the same $K$ on 2D
Mat\'ern-$\nicefrac{3}{2}$ (batch$=1$, $N_{\text{samples}}{=}1$).
Radix sort runs in $O(wK)$ time on $w$-bit keys, and at curve
order $h{=}16$ the keys are 32-bit unsigned integers with each 2D
point quantised to one of $2^{32}$ lattice
cells~\citep{sagan1994sfc, hamilton2008chi}, so $w{=}32$ is fixed by
the curve order rather than by $K$; the sort therefore has $O(K)$
complexity within any regime in which Hilbert keys remain distinct.
Once $K$ grows large enough to force lattice collisions the sort
must break ties arbitrarily and the locality argument
of~\citet{moon2001hilbert} no longer applies. Above $K \gtrsim
65{,}000$ the compiled program lowers \texttt{argsort} to CUB's
radix-sort kernel~\citep{merrill2016cub}; below this threshold an
inline path performs a fixed number of comparisons per element.
Both regimes are $O(K)$. All measurements are on a single NVIDIA L40S
in one Python process with one JAX trace cache, $10$ warm-ups
followed by $100$ timed reps, median statistic.

The identity baseline is constant at ${\sim}40\,\mu$s across four
orders of magnitude in $K$. Kernel-only cost grows from $22.1\,\mu$s
at $K{=}2{,}048$ to $83.6\,\mu$s at $K{=}10^{6}$, a $3.8{\times}$
increase for a $512{\times}$ increase in $K$. Above
$K \approx 5{\times}10^{5}$ the radix kernel saturates memory
bandwidth and per-element cost flattens. The sort fraction of MFNP
inference plateaus in the $1.0$--$1.9\%$ range for $K \leq 8{,}192$
where both kernels are dispatch-bound, then decreases monotonically
with $K$ to $0.014\%$ at $K{=}10^{6}$ as the linear-time scan
dominates.

\begin{table}[!htbp]
\centering
\caption{Sort scaling and its share of MFNP inference (NVIDIA L40S, JAX 0.9.0.1, CUDA 12). Identity is a same-shape no-op baseline; kernel is sort total minus identity. MFNP wall-clock is per-forward at the same $K$ on 2D Mat\'ern-$\nicefrac{3}{2}$ (1-NFE Euler, batch$=1$, $N_{\text{samples}}{=}1$, headline checkpoint).}
\label{tab:sort-scaling}
\scriptsize
\setlength{\tabcolsep}{6pt}
\begin{tabular}{@{} r r r r r r @{}}
\toprule
 & \multicolumn{3}{c}{\textit{Standalone sort}} & \multicolumn{2}{c}{\textit{Share of MFNP inference}} \\
\cmidrule(lr){2-4} \cmidrule(lr){5-6}
$K$ & \textbf{Identity ($\mu$s)} & \textbf{Sort ($\mu$s)} & \textbf{Kernel ($\mu$s)} & \textbf{MFNP total (ms)} & \textbf{Sort fraction} \\
\midrule
$128$           & $41.8$ & $44.4$ & $2.6$ & $3.41$ & $1.30\%$ \\
$256$           & $37.9$ & $35.4$ & ${\approx}0$ & $3.50$ & $1.01\%$ \\
$512$           & $41.3$ & $45.3$ & $4.0$ & $3.61$ & $1.26\%$ \\
$1{,}024$       & $41.4$ & $44.2$ & $2.7$ & $3.72$ & $1.19\%$ \\
$2{,}048$       & $35.9$ & $58.0$ & $22.1$ & $3.87$ & $1.50\%$ \\
$4{,}096$       & $36.5$ & $86.7$ & $50.3$ & $4.67$ & $1.86\%$ \\
$8{,}192$       & $35.9$ & $86.1$ & $50.2$ & $7.27$ & $1.18\%$ \\
$16{,}384$      & $38.6$ & $101.7$ & $63.1$ & $14.02$ & $0.73\%$ \\
$32{,}768$      & $36.6$ & $102.4$ & $65.8$ & $31.76$ & $0.32\%$ \\
$65{,}536$      & $31.8$ & $97.7$ & $66.0$ & $61.96$ & $0.16\%$ \\
$131{,}072$     & $40.9$ & $101.4$ & $60.5$ & $120.23$ & $0.084\%$ \\
$262{,}144$     & $36.3$ & $101.2$ & $65.0$ & $233.14$ & $0.043\%$ \\
$524{,}288$     & $40.7$ & $115.2$ & $74.5$ & $460.38$ & $0.025\%$ \\
$1{,}048{,}576$ & $41.8$ & $125.4$ & $83.6$ & $912.48$ & $0.014\%$ \\
\bottomrule
\end{tabular}
\end{table}

\section{Full matched-regime results}
\label{app:synth-full}

Table~\ref{tab:synth-full} covers the full 1D sweep at $K{=}128$ across RBF, Mat\'ern-$\nicefrac{3}{2}$, and Mat\'ern-$\nicefrac{5}{2}$; the 2D sweep is in the main-body Table~\ref{tab:synth-main}.

\begin{table}[!htbp]
\centering
\caption{Full 1D matched-regime results at $K_{\max}{=}128$, all three kernels (RBF, Mat\'ern-$\nicefrac{3}{2}$, Mat\'ern-$\nicefrac{5}{2}$). Baselines are grouped by predictive class into \emph{factorised} ($O(K)$ pooling or $O(NM)$ cross-attention with per-query Gaussian head), \emph{joint / quadratic} ($O(K^2)$ self-attention), \emph{joint / linear} ($O(K)$ Mamba-2 scan). MSE, CRPS ($\downarrow$); C2ST (target $0.5$); $95\%$ coverage (target nominal); time = per-episode ms, MFNP and baselines on one NVIDIA L40S. Mean $\pm$ SE over 10 evaluation seeds; bold = best per column (within 1 SE). \textsuperscript{\dag} our method.}
\label{tab:synth-full}
\scriptsize
\setlength{\tabcolsep}{4pt}
\begin{tabular}{@{} l c c c c c @{}}
\toprule
 & \multicolumn{2}{c}{\textbf{Predictive Quality}} & \multicolumn{2}{c}{\textbf{Calibration Quality}} & \\
\cmidrule(lr){2-3} \cmidrule(lr){4-5}
\textbf{Model} & \textbf{MSE}\,$\downarrow$ & \textbf{CRPS}\,$\downarrow$ & \textbf{C2ST}\,$\rightarrow\!0.5$ & \textbf{Cov-95}\,$\rightarrow\!0.95$ & \textbf{Time\,(ms)}\,$\downarrow$ \\
\midrule

\multicolumn{6}{l}{\textit{1D Gaussian -- RBF} ($K{=}128$)} \\[2pt]
\multicolumn{6}{@{}l}{\;\; \textit{Factorised predictives}} \\
  CNP  & $0.0285\pm0.0008$ & $0.0666\pm0.0009$ & $0.673\pm0.020$ & $0.941\pm0.002$ & $\mathbf{0.40\pm0.02}$ \\
  NP   & $0.0388\pm0.0010$ & $0.0906\pm0.0014$ & $0.753\pm0.029$ & $0.474\pm0.003$ & $0.58\pm0.00$ \\
  ANP  & $0.0026\pm0.0001$ & $0.0255\pm0.0002$ & $0.691\pm0.008$ & $0.091\pm0.001$ & $1.31\pm0.01$ \\
  CANP & $0.0047\pm0.0002$ & $0.0252\pm0.0003$ & $0.605\pm0.005$ & $0.880\pm0.001$ & $1.09\pm0.01$ \\
\multicolumn{6}{@{}l}{\;\; \textit{Joint predictives, quadratic-time}} \\
  TNPA & $0.0029\pm0.0002$ & $0.0197\pm0.0002$ & $\mathbf{0.545\pm0.008}$ & $\mathbf{0.956\pm0.001}$ & $442.9\pm0.7$ \\
  NDP  & $0.2461\pm0.0047$ & $0.2772\pm0.0023$ & $0.742\pm0.026$ & $0.966\pm0.001$ & $929.0\pm6.5$ \\
  FNP  & $\mathbf{0.0018\pm0.0001}$ & $\mathbf{0.0180\pm0.0001}$ & $\mathbf{0.540\pm0.014}$ & $0.664\pm0.002$ & $(1.516\pm0.002){\times}10^{3}$ \\
\multicolumn{6}{@{}l}{\;\; \textit{Joint predictives, linear-time}} \\
  MFNP\textsuperscript{\dag} & $0.0034\pm0.0002$ & $0.0224\pm0.0002$ & $0.596\pm0.006$ & $0.863\pm0.001$ & $25.7\pm0.2$ \\
\addlinespace

\midrule

\multicolumn{6}{l}{\textit{1D Gaussian -- Mat\'ern-$\nicefrac{3}{2}$} ($K{=}128$)} \\[2pt]
\multicolumn{6}{@{}l}{\;\; \textit{Factorised predictives}} \\
  CNP  & $0.0556\pm0.0011$ & $0.1085\pm0.0010$ & $0.695\pm0.014$ & $0.931\pm0.001$ & $\mathbf{0.47\pm0.02}$ \\
  NP   & $0.0466\pm0.0007$ & $0.1139\pm0.0010$ & $0.756\pm0.019$ & $0.505\pm0.002$ & $0.62\pm0.03$ \\
  ANP  & $0.0111\pm0.0003$ & $0.0504\pm0.0005$ & $0.643\pm0.006$ & $0.148\pm0.001$ & $1.12\pm0.01$ \\
  CANP & $0.0156\pm0.0003$ & $0.0512\pm0.0005$ & $0.593\pm0.009$ & $0.856\pm0.002$ & $1.02\pm0.01$ \\
\multicolumn{6}{@{}l}{\;\; \textit{Joint predictives, quadratic-time}} \\
  TNPA & $\mathbf{0.0110\pm0.0003}$ & $\mathbf{0.0367\pm0.0004}$ & $\mathbf{0.499\pm0.003}$ & $\mathbf{0.946\pm0.001}$ & $735.7\pm0.4$ \\
  NDP  & $0.1720\pm0.0028$ & $0.2508\pm0.0014$ & $0.688\pm0.018$ & $0.996\pm0.000$ & $(2.174\pm0.004){\times}10^{3}$ \\
  FNP  & $\mathbf{0.0107\pm0.0002}$ & $0.0398\pm0.0004$ & $0.532\pm0.011$ & $0.752\pm0.001$ & $(1.253\pm0.001){\times}10^{3}$ \\
\multicolumn{6}{@{}l}{\;\; \textit{Joint predictives, linear-time}} \\
  MFNP\textsuperscript{\dag} & $0.0111\pm0.0003$ & $0.0398\pm0.0004$ & $0.537\pm0.006$ & $0.938\pm0.001$ & $29.5\pm0.2$ \\
\addlinespace

\midrule

\multicolumn{6}{l}{\textit{1D Gaussian -- Mat\'ern-$\nicefrac{5}{2}$} ($K{=}128$)} \\[2pt]
\multicolumn{6}{@{}l}{\;\; \textit{Factorised predictives}} \\
  CNP  & $0.0218\pm0.0005$ & $0.0595\pm0.0006$ & $0.657\pm0.017$ & $0.930\pm0.001$ & $\mathbf{0.49\pm0.02}$ \\
  NP   & $0.0400\pm0.0009$ & $0.1070\pm0.0013$ & $0.815\pm0.023$ & $0.333\pm0.002$ & $0.53\pm0.00$ \\
  ANP  & $0.0064\pm0.0002$ & $0.0367\pm0.0004$ & $0.661\pm0.004$ & $0.131\pm0.001$ & $1.18\pm0.01$ \\
  CANP & $0.0077\pm0.0002$ & $0.0320\pm0.0003$ & $0.577\pm0.005$ & $0.883\pm0.001$ & $1.05\pm0.01$ \\
\multicolumn{6}{@{}l}{\;\; \textit{Joint predictives, quadratic-time}} \\
  TNPA & $\mathbf{0.0058\pm0.0002}$ & $\mathbf{0.0269\pm0.0003}$ & $\mathbf{0.512\pm0.008}$ & $\mathbf{0.953\pm0.001}$ & $(1.601\pm0.002){\times}10^{3}$ \\
  NDP  & $0.0745\pm0.0010$ & $0.1441\pm0.0009$ & $0.737\pm0.008$ & $0.977\pm0.001$ & $(2.779\pm0.005){\times}10^{3}$ \\
  FNP  & $\mathbf{0.0057\pm0.0002}$ & $0.0279\pm0.0003$ & $0.531\pm0.010$ & $0.680\pm0.002$ & $534.2\pm0.8$ \\
\multicolumn{6}{@{}l}{\;\; \textit{Joint predictives, linear-time}} \\
  MFNP\textsuperscript{\dag} & $0.0065\pm0.0002$ & $0.0303\pm0.0003$ & $0.541\pm0.007$ & $0.970\pm0.001$ & $8.4\pm0.1$ \\
\bottomrule
\end{tabular}
\end{table}

\section{Scaling}
\label{app:scaling}

Table~\ref{tab:scaling} reports the end-to-end sampling latency plotted in Figure~\ref{fig:scaling} for each joint-predictive NP against total sequence length $K$ on 2D Gaussian Mat\'ern-$\nicefrac{3}{2}$.

\begin{table}[!htbp]
\centering
\caption{End-to-end sampling latency (ms) for joint-predictive NPs vs.\ total sequence length $K$ on 2D Gaussian Mat\'ern-$\nicefrac{3}{2}$ on one NVIDIA L40S, batch$=1$, $N_{\text{samples}}{=}1$ (one full posterior draw per pass, with 1-NFE Euler for MFNP and the published step count for each baseline). Median of $5$ reps after $3$ warm-ups; relative SE $<5\%$ on every completing cell. Values $\geq 10^{3}$ in scientific notation, rounded to $2$--$3$ sig figs. \textsc{oom} = out of memory. Bold = best per column. \textsuperscript{\dag} our method.}
\label{tab:scaling}
\scriptsize
\setlength{\tabcolsep}{6pt}
\begin{tabular}{@{} l r r r r r r @{}}
\toprule
\textbf{Model} & $16$ & $256$ & $4{,}096$ & $16{,}384$ & $65{,}536$ & $1{,}048{,}576$ \\
\midrule
\multicolumn{7}{@{}l}{\;\; \textit{Joint predictives, quadratic-time}} \\
  TNPA   & $10.7$ & $151$ & $1.2{\times}10^{5}$ & $7.4{\times}10^{6}$ & \textsc{oom} & \textsc{oom} \\
  NDP    & $315$  & $317$ & $1.5{\times}10^{3}$ & \textsc{oom} & \textsc{oom} & \textsc{oom} \\
  FNP    & $124.0$ & $122.6$ & $1.07{\times}10^{4}$ & \textsc{oom} & \textsc{oom} & \textsc{oom} \\
\multicolumn{7}{@{}l}{\;\; \textit{Joint predictives, linear-time}} \\
  MFNP\textsuperscript{\dag} & $\mathbf{3.51}$ & $\mathbf{3.63}$ & $\mathbf{4.82}$ & $\mathbf{14.74}$ & $\mathbf{61.88}$ & $\mathbf{913.86}$ \\
\bottomrule
\end{tabular}
\end{table}

\section{Full NOAA results}
\label{app:full-noaa}

\begin{figure}[!htbp]
\centering
\includegraphics[width=\linewidth]{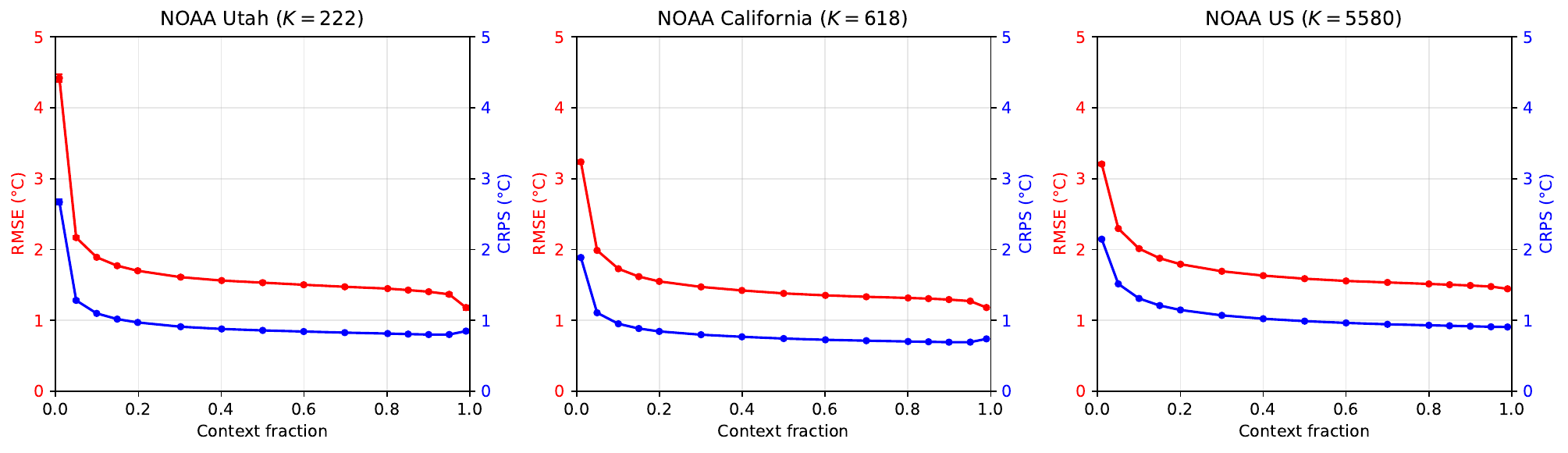}
\caption{MFNP predictive accuracy as a function of context fraction $N/K$ on the three NOAA tiers (Utah $K{=}222$; California $K{=}486$; continental US $K{=}5{,}580$), averaged over $1{,}024$ random (date, station-partition) draws per fraction. Red (left axis) shows RMSE and blue (right axis) shows CRPS, both in \textdegree C.}
\label{fig:noaa-ranges}
\end{figure}

\paragraph{Context-fraction sweep.}
Figure~\ref{fig:noaa-ranges} sweeps the context fraction $N/K$ across fifteen values from $1\%$ to $99\%$ on each NOAA tier, holding $N+M$ fixed at the full station count. Each fraction is averaged over $1{,}024$ random (date, station-partition) draws, so the curves report the population-level dependence of MFNP error on the size of the observed support rather than a single realisation. RMSE and CRPS fall steeply over the first $\sim 30\%$ of context (roughly $3{\times}$ on Utah, from $4.9$ to $1.6$\textdegree C RMSE, and $2{\times}$ on California and the US) and then flatten out, with the remaining context yielding a further $12$--$21\%$ of RMSE reduction. The $50\%$ split used for the headline tables (Table~\ref{tab:noaa-full}) sits in this plateau.

Table~\ref{tab:noaa-full} covers the full NOAA sweep, adding the factorised baselines (CNP, NP, ANP, CANP) at each of the three tiers.

\begin{table}[!htbp]
\centering
\caption{Full NOAA temperature interpolation on Utah ($K{=}222$), California ($K{=}486$), and the continental US ($K{=}5{,}580$); baselines grouped by predictive class as in Table~\ref{tab:synth-full}. MSE (\textdegree C)$^2$, RMSE, CRPS \textdegree C ($\downarrow$); Cov-95 (target $0.95$); C2ST omitted (no tractable reference posterior on real fields); time = per-episode ms on one NVIDIA L40S. Mean $\pm$ SE over 10 evaluation seeds, denormalised scale; bold = best per column within tier (within 1 SE). \textsuperscript{\dag} our method.}
\label{tab:noaa-full}
\scriptsize
\setlength{\tabcolsep}{4pt}
\begin{tabular}{@{} l c c c c c @{}}
\toprule
 & \multicolumn{3}{c}{\textbf{Predictive Quality}} & \textbf{Calibration} & \\
\cmidrule(lr){2-4} \cmidrule(lr){5-5}
\textbf{Model} & \textbf{MSE (\textdegree C)$^2$\,$\downarrow$} & \textbf{RMSE (\textdegree C)\,$\downarrow$} & \textbf{CRPS (\textdegree C)\,$\downarrow$} & \textbf{Cov-95}\,$\rightarrow\!0.95$ & \textbf{Time\,(ms)}\,$\downarrow$ \\
\midrule

\multicolumn{6}{l}{\textit{NOAA Utah} ($K{=}222$)} \\[2pt]
\multicolumn{6}{@{}l}{\;\; \textit{Factorised predictives}} \\
  CNP  & $4.230\pm0.010$ & $2.057\pm0.002$ & $1.075\pm0.001$ & $0.930\pm0.000$ & $\mathbf{0.46\pm0.02}$ \\
  NP   & $6.196\pm0.017$ & $2.489\pm0.003$ & $1.551\pm0.002$ & $0.481\pm0.000$ & $0.75\pm0.00$ \\
  ANP  & $3.653\pm0.010$ & $1.911\pm0.003$ & $1.246\pm0.002$ & $0.268\pm0.000$ & $1.34\pm0.06$ \\
  CANP & $3.783\pm0.011$ & $1.945\pm0.003$ & $1.005\pm0.001$ & $0.900\pm0.000$ & $0.99\pm0.01$ \\
\multicolumn{6}{@{}l}{\;\; \textit{Joint predictives, quadratic-time}} \\
  TNPA & $2.524\pm0.007$ & $1.589\pm0.002$ & $0.783\pm0.001$ & $0.846\pm0.000$ & $(4.923\pm0.004){\times}10^{3}$ \\
  NDP  & $4.665\pm0.012$ & $2.160\pm0.003$ & $1.221\pm0.002$ & $0.742\pm0.000$ & $(2.414\pm0.004){\times}10^{3}$ \\
  FNP  & $2.712\pm0.007$ & $1.647\pm0.002$ & $0.853\pm0.001$ & $0.730\pm0.000$ & $(3.071\pm0.003){\times}10^{3}$ \\
\multicolumn{6}{@{}l}{\;\; \textit{Joint predictives, linear-time}} \\
  MFNP\textsuperscript{\dag} & $\mathbf{2.464\pm0.016}$ & $\mathbf{1.570\pm0.005}$ & $\mathbf{0.767\pm0.002}$ & $\mathbf{0.940\pm0.000}$ & $43.3\pm0.4$ \\
\addlinespace

\midrule

\multicolumn{6}{l}{\textit{NOAA California} ($K{=}486$)} \\[2pt]
\multicolumn{6}{@{}l}{\;\; \textit{Factorised predictives}} \\
  CNP  & $3.575\pm0.015$ & $1.891\pm0.004$ & $1.013\pm0.002$ & $0.938\pm0.000$ & $\mathbf{0.48\pm0.02}$ \\
  NP   & $4.365\pm0.021$ & $2.089\pm0.005$ & $1.323\pm0.003$ & $0.496\pm0.001$ & $1.24\pm0.00$ \\
  ANP  & $3.340\pm0.012$ & $1.828\pm0.003$ & $1.213\pm0.002$ & $0.331\pm0.000$ & $2.35\pm0.01$ \\
  CANP & $3.431\pm0.015$ & $1.852\pm0.004$ & $0.976\pm0.002$ & $0.892\pm0.000$ & $1.08\pm0.05$ \\
\multicolumn{6}{@{}l}{\;\; \textit{Joint predictives, quadratic-time}} \\
  TNPA & $1.898\pm0.005$ & $1.378\pm0.002$ & $0.701\pm0.001$ & $0.938\pm0.000$ & $(3.004\pm0.002){\times}10^{4}$ \\
  NDP  & $3.071\pm0.008$ & $1.753\pm0.002$ & $1.064\pm0.001$ & $0.590\pm0.000$ & $(9.154\pm0.015){\times}10^{3}$ \\
  FNP  & $2.395\pm0.008$ & $1.548\pm0.003$ & $0.839\pm0.001$ & $0.785\pm0.001$ & $(8.844\pm0.004){\times}10^{3}$ \\
\multicolumn{6}{@{}l}{\;\; \textit{Joint predictives, linear-time}} \\
  MFNP\textsuperscript{\dag} & $\mathbf{1.808\pm0.006}$ & $\mathbf{1.344\pm0.002}$ & $\mathbf{0.681\pm0.001}$ & $\mathbf{0.945\pm0.000}$ & $67.9\pm0.5$ \\
\addlinespace

\midrule

\multicolumn{6}{l}{\textit{NOAA US} ($K{=}5{,}580$)} \\[2pt]
\multicolumn{6}{@{}l}{\;\; \textit{Factorised predictives}} \\
  CNP  & $7.764\pm0.032$ & $2.786\pm0.006$ & $1.501\pm0.003$ & $0.911\pm0.000$ & $\mathbf{2.35\pm0.69}$ \\
  NP   & $11.039\pm0.034$ & $3.322\pm0.005$ & $2.349\pm0.003$ & $0.229\pm0.000$ & $33.5\pm0.0$ \\
  ANP  & $9.312\pm0.025$ & $3.052\pm0.004$ & $1.999\pm0.002$ & $0.272\pm0.000$ & $25.9\pm0.0$ \\
  CANP & $5.585\pm0.027$ & $2.363\pm0.006$ & $1.196\pm0.002$ & $0.905\pm0.000$ & $8.45\pm0.01$ \\
\multicolumn{6}{@{}l}{\;\; \textit{Joint predictives, quadratic-time}} \\
  TNPA & \textsc{t/o}     & \textsc{t/o}    & \textsc{t/o}    & \textsc{t/o} & \textsc{t/o} \\
  NDP  & $3.528\pm0.087$   & $1.878\pm0.023$  & $1.194\pm0.013$  & $0.322\pm0.005$ & $(7.075\pm0.020){\times}10^{5}$ \\
  FNP  & $2.872\pm0.064$& $1.695\pm0.019$  & $0.864\pm0.009$  & $0.786\pm0.001$ & $(8.831\pm0.002){\times}10^{5}$ \\
\multicolumn{6}{@{}l}{\;\; \textit{Joint predictives, linear-time}} \\
  MFNP\textsuperscript{\dag} & $\mathbf{2.554\pm0.010}$ & $\mathbf{1.598\pm0.003}$ & $\mathbf{0.774\pm0.001}$ & $\mathbf{0.955\pm0.001}$ & $(1.126\pm0.000){\times}10^{3}$ \\
\bottomrule
\end{tabular}
\end{table}

\paragraph{Gaussian-process references.}
Table~\ref{tab:noaa-gp} reports three GPs on the NOAA test sets. Each
uses an ARD Mat\'ern-$\nicefrac{3}{2}$ kernel on the station
coordinates and fits its hyperparameters to the context of each
episode. The exact GP solves the full $N \times N$ system, SGPR uses
inducing points~\citep{titsias2009variational}, and the
nearest-neighbour GP conditions each point on its $20$ nearest
neighbours~\citep{vecchia1988estimation, datta2016hierarchical}. MFNP
is trained across dates on the same station network and the GPs use
only the current episode.

\begin{table}[!htbp]
\centering
\caption{Gaussian-process references on NOAA. MSE (\textdegree C)$^2$, RMSE, CRPS \textdegree C ($\downarrow$); Cov-95 (target $0.95$). Mean $\pm$ SE over evaluation seeds ($10$ for MFNP and the nearest-neighbour GP; $10$, $10$ and $8$ for the exact GP; $4$, $4$ and $3$ for SGPR). SGPR uses $64$ inducing points on Utah and California and $512$ on the US. \textsuperscript{\dag} our method.}
\label{tab:noaa-gp}
\scriptsize
\setlength{\tabcolsep}{4pt}
\begin{tabular}{@{} l c c c c @{}}
\toprule
\textbf{Model} & \textbf{MSE (\textdegree C)$^2$\,$\downarrow$} & \textbf{RMSE (\textdegree C)\,$\downarrow$} & \textbf{CRPS (\textdegree C)\,$\downarrow$} & \textbf{Cov-95}\,$\rightarrow\!0.95$ \\
\midrule
\multicolumn{5}{l}{\textit{NOAA Utah} ($K{=}222$)} \\[2pt]
  Exact GP              & $16.240\pm0.034$ & $3.936\pm0.004$ & $2.246\pm0.002$ & $0.916\pm0.000$ \\
  SGPR                  & $16.501\pm0.073$ & $3.965\pm0.007$ & $2.277\pm0.004$ & $0.936\pm0.001$ \\
  Nearest-neighbour GP  & $16.255\pm0.065$ & $3.937\pm0.007$ & $2.239\pm0.004$ & $0.916\pm0.000$ \\
  MFNP\textsuperscript{\dag} & $\mathbf{2.464\pm0.016}$ & $\mathbf{1.570\pm0.005}$ & $\mathbf{0.767\pm0.002}$ & $\mathbf{0.940\pm0.000}$ \\
\midrule
\multicolumn{5}{l}{\textit{NOAA California} ($K{=}486$)} \\[2pt]
  Exact GP              & $11.813\pm0.049$ & $3.399\pm0.007$ & $1.896\pm0.004$ & $0.924\pm0.000$ \\
  SGPR                  & $12.330\pm0.057$ & $3.476\pm0.008$ & $1.948\pm0.004$ & $0.936\pm0.000$ \\
  Nearest-neighbour GP  & $12.119\pm0.050$ & $3.441\pm0.007$ & $1.914\pm0.004$ & $0.921\pm0.000$ \\
  MFNP\textsuperscript{\dag} & $\mathbf{1.808\pm0.006}$ & $\mathbf{1.344\pm0.002}$ & $\mathbf{0.681\pm0.001}$ & $\mathbf{0.945\pm0.000}$ \\
\midrule
\multicolumn{5}{l}{\textit{NOAA US} ($K{=}5{,}580$)} \\[2pt]
  Exact GP              & $9.440\pm0.025$ & $3.050\pm0.004$ & $1.690\pm0.002$ & $0.929\pm0.000$ \\
  SGPR                  & $9.757\pm0.009$ & $3.101\pm0.001$ & $1.724\pm0.001$ & $0.932\pm0.000$ \\
  Nearest-neighbour GP  & $9.498\pm0.032$ & $3.059\pm0.005$ & $1.692\pm0.003$ & $0.925\pm0.000$ \\
  MFNP\textsuperscript{\dag} & $\mathbf{2.554\pm0.010}$ & $\mathbf{1.598\pm0.003}$ & $\mathbf{0.774\pm0.001}$ & $\mathbf{0.955\pm0.001}$ \\
\bottomrule
\end{tabular}
\end{table}

\begin{figure}[!htbp]
\centering
\includegraphics[width=\linewidth]{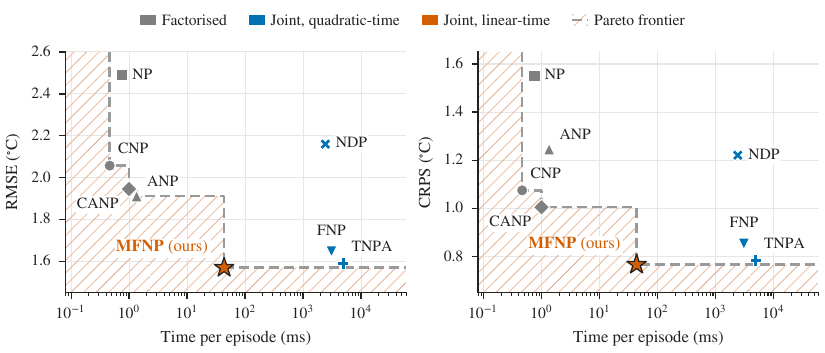}
\caption{Per-episode latency (x, log) versus RMSE (left panel) and CRPS (right panel) on NOAA Utah ($K{=}222$). Conventions as in Figure~\ref{fig:noaa-pareto}.}
\label{fig:noaa-pareto-utah}
\end{figure}

\begin{figure}[!htbp]
\centering
\includegraphics[width=\linewidth]{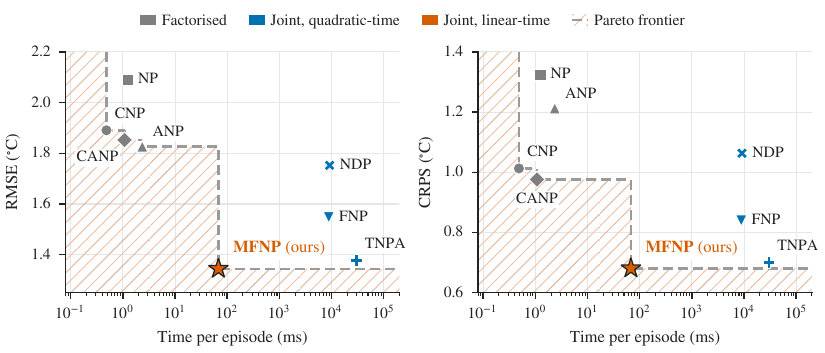}
\caption{Per-episode latency (x, log) versus RMSE (left panel) and CRPS (right panel) on NOAA California ($K{=}486$). Conventions as in Figure~\ref{fig:noaa-pareto}.}
\label{fig:noaa-pareto-california}
\end{figure}

\paragraph{Per-tier Pareto frontiers.}
Figures~\ref{fig:noaa-pareto-utah} and~\ref{fig:noaa-pareto-california} reproduce the RMSE--latency analysis of Figure~\ref{fig:noaa-pareto} on the Utah and California tiers; marker, error-bar, and Pareto-staircase conventions are identical.

\begin{figure}[!htbp]
\centering
\includegraphics[width=\linewidth]{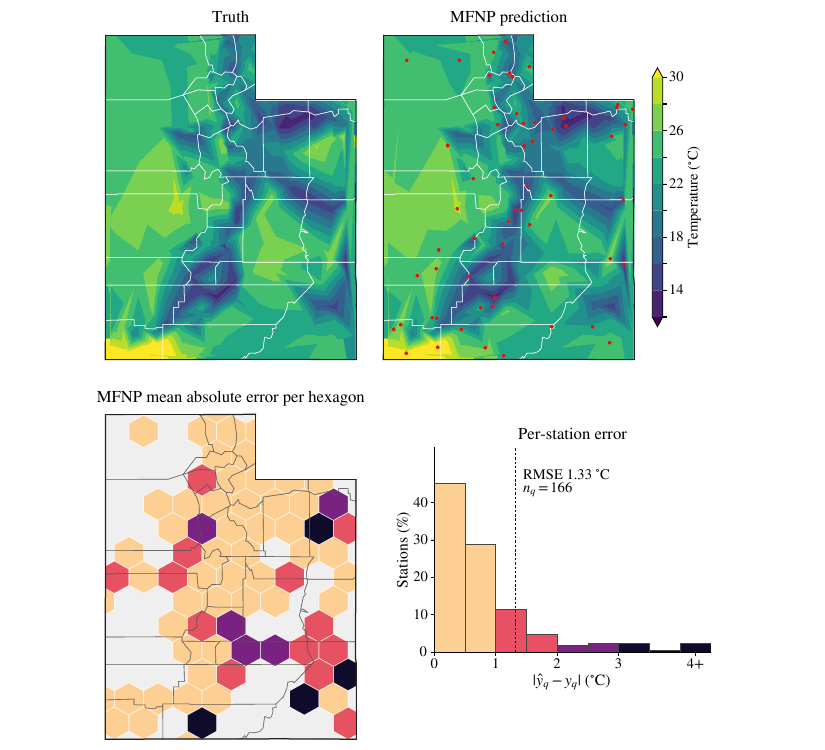}
\caption{NOAA Utah ($K{=}222$) posterior on a held-out date, $25\%/75\%$ context/query split. Panels as in Figure~\ref{fig:noaa-qual-us}, with ${\sim}50$\,km hexagons.}
\label{fig:noaa-qual-utah}
\end{figure}

\begin{figure}[!htbp]
\centering
\includegraphics[width=\linewidth]{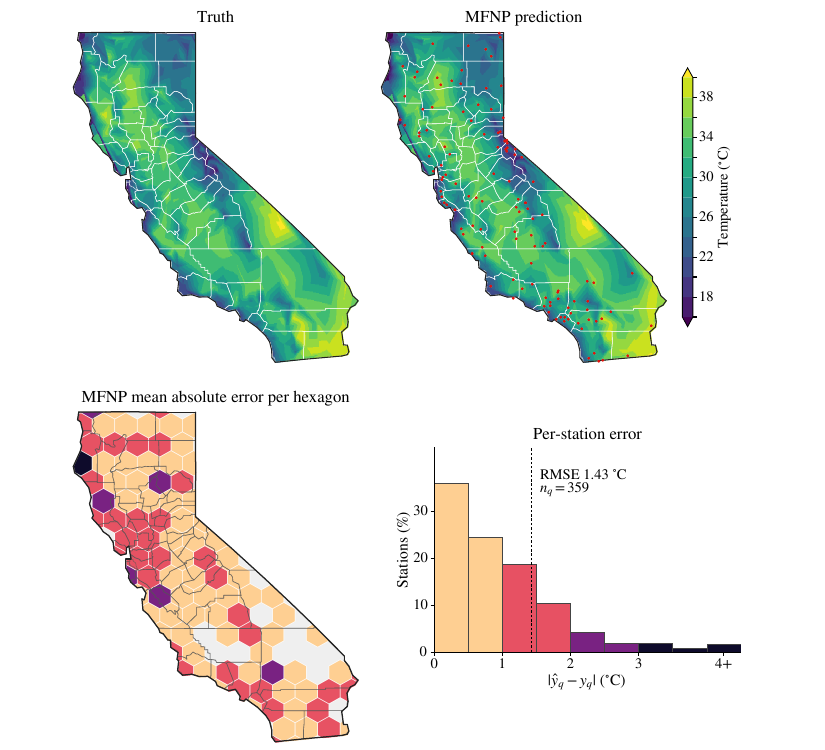}
\caption{NOAA California ($K{=}486$) posterior on a held-out date, $25\%/75\%$ context/query split. Panels as in Figure~\ref{fig:noaa-qual-us}, with ${\sim}70$\,km hexagons. Five coastal stations outside the state outline are omitted.}
\label{fig:noaa-qual-california}
\end{figure}

\paragraph{Qualitative posteriors.}
Figures~\ref{fig:noaa-qual-utah}--\ref{fig:noaa-qual-california} show MFNP posterior predictions on a single held-out date for Utah and California at a $25\%/75\%$ context/query split, in the layout of Figure~\ref{fig:noaa-qual-us}. They show the predicted field and the spatial structure of its error on a single realisation. The population-averaged dependence on context fraction is reported separately in Figure~\ref{fig:noaa-ranges}.

\section{Architectural ablations}
\label{app:ablations}

Table~\ref{tab:ablation} reports the architectural ablations. Each row retrains MFNP from scratch under the full hyperparameter configuration and uses the same evaluation protocol as the main tables ($N_{\text{samples}}{=}128$, batch~$=1$, 10 evaluation seeds). The full row matches the headline 2D synthetic and NOAA configurations.

\textbf{Variants.} \emph{Scan routing.} \emph{Single direction} uses only $\pi_1$; \emph{two directions} uses $\pi_1$ and its reversal (no axis-transposed curve). In 1D there is no transposed curve, so the full model fills its four direction slots with the ascending and descending orders twice, and \emph{two directions} uses each order once. \emph{Point ordering.} \emph{Random order} uses a random permutation per episode, isolating the effect of locality in the serialisation. \emph{State flow.} \emph{Single hidden state} runs a unidirectional SSM over the context, takes its final hidden state as a single summary, and initialises a separate decoder over queries from that vector.

\begin{table}[!htbp]
\centering
\caption{Architectural ablations on MambaFlowNP. MSE/CRPS on each task's native scale (dimensionless synthetic; (\textdegree C)$^2$/\textdegree C NOAA). Mean $\pm$ SE over 10 evaluation seeds at $N_{\text{samples}}{=}128$ on a single training checkpoint per variant; bold = best per column within block (within 1 SE); times on one NVIDIA L40S. C2ST omitted on NOAA tiers (no tractable reference posterior on real fields), shown as \texttt{--}. \textsuperscript{\dag} our method.}
\label{tab:ablation}
\scriptsize
\setlength{\tabcolsep}{4pt}
\begin{tabular}{@{} l c c c c c @{}}
\toprule
 & \multicolumn{2}{c}{\textbf{Predictive Quality}} & \multicolumn{2}{c}{\textbf{Calibration Quality}} & \\
\cmidrule(lr){2-3} \cmidrule(lr){4-5}
\textbf{Variant} & \textbf{MSE}\,$\downarrow$ & \textbf{CRPS}\,$\downarrow$ & \textbf{C2ST}\,$\rightarrow\!0.5$ & \textbf{Cov-95}\,$\rightarrow\!0.95$ & \textbf{Time\,(ms)}\,$\downarrow$ \\
\midrule

\multicolumn{6}{l}{\textit{1D Gaussian -- Mat\'ern-$\nicefrac{3}{2}$} ($K{=}128$)} \\[2pt]
  Single direction          & $0.0471\pm0.0012$ & $0.0860\pm0.0010$ & $0.661\pm0.012$ & $0.893\pm0.002$ & $\mathbf{8.0\pm0.1}$ \\
  Two directions            & $\mathbf{0.0101\pm0.0002}$ & $\mathbf{0.0373\pm0.0004}$ & $\mathbf{0.515\pm0.003}$ & $0.936\pm0.001$ & $13.3\pm0.1$ \\
  Random order              & $0.0191\pm0.0004$ & $0.0552\pm0.0005$ & $0.619\pm0.017$ & $0.922\pm0.001$ & $28.8\pm0.3$ \\
  Single hidden state       & $0.0122\pm0.0002$ & $0.0425\pm0.0004$ & $0.562\pm0.007$ & $0.926\pm0.001$ & $10.1\pm0.5$ \\
  MFNP\textsuperscript{\dag} (full, 2 dir.\ $\times 2$) & $0.0111\pm0.0003$ & $0.0398\pm0.0004$ & $0.537\pm0.006$ & $\mathbf{0.938\pm0.001}$ & $29.5\pm0.2$ \\
\midrule

\multicolumn{6}{l}{\textit{2D Gaussian -- Mat\'ern-$\nicefrac{3}{2}$} ($K{=}256$)} \\[2pt]
  Single direction          & $0.0281\pm0.0008$ & $0.0709\pm0.0008$ & $0.584\pm0.008$ & $0.989\pm0.000$ & $\mathbf{25.2\pm0.2}$ \\
  Two directions            & $\mathbf{0.0163\pm0.0005}$ & $\mathbf{0.0537\pm0.0006}$ & $0.575\pm0.005$ & $0.856\pm0.001$ & $52.1\pm0.3$ \\
  Random order              & $0.0197\pm0.0006$ & $0.0596\pm0.0006$ & $0.588\pm0.007$ & $\mathbf{0.962\pm0.001}$ & $103.3\pm0.3$ \\
  Single hidden state       & $0.0196\pm0.0005$ & $0.0608\pm0.0006$ & $0.612\pm0.009$ & $0.885\pm0.000$ & $25.8\pm0.2$ \\
  MFNP\textsuperscript{\dag} (full, 4-dir) & $\mathbf{0.0167\pm0.0005}$ & $\mathbf{0.0537\pm0.0005}$ & $\mathbf{0.554\pm0.005}$ & $0.971\pm0.000$ & $103.6\pm0.3$ \\
\midrule

\multicolumn{6}{l}{\textit{NOAA Utah} ($K{=}222$)} \\[2pt]
  Single direction          & $4.055\pm0.008$ & $0.996\pm0.001$ & -- & $0.714\pm0.000$ & $6.04\pm0.07$ \\
  Two directions            & $2.528\pm0.007$ & $0.852\pm0.001$ & -- & $0.698\pm0.000$ & $13.44\pm0.26$ \\
  Random order              & $4.940\pm0.013$ & $1.272\pm0.002$ & -- & $0.645\pm0.000$ & $27.89\pm0.10$ \\
  Single hidden state       & $4.060\pm0.008$ & $1.007\pm0.001$ & -- & $0.682\pm0.000$ & $\mathbf{5.85\pm0.01}$ \\
  MFNP\textsuperscript{\dag} (full, 4-dir) & $\mathbf{2.464\pm0.016}$ & $\mathbf{0.767\pm0.002}$ & -- & $\mathbf{0.940\pm0.000}$ & $43.3\pm0.4$ \\
\midrule

\multicolumn{6}{l}{\textit{NOAA California} ($K{=}486$)} \\[2pt]
  Single direction          & $2.860\pm0.006$ & $0.858\pm0.001$ & -- & $0.840\pm0.000$ & $\mathbf{25.32\pm0.10}$ \\
  Two directions            & $1.965\pm0.006$ & $0.748\pm0.001$ & -- & $0.830\pm0.000$ & $50.52\pm0.13$ \\
  Random order              & $9.111\pm0.049$ & $1.755\pm0.004$ & -- & $0.935\pm0.000$ & $96.98\pm0.11$ \\
  Single hidden state       & $2.855\pm0.006$ & $0.861\pm0.001$ & -- & $0.832\pm0.000$ & $\mathbf{25.22\pm0.08}$ \\
  MFNP\textsuperscript{\dag} (full, 4-dir) & $\mathbf{1.808\pm0.006}$ & $\mathbf{0.681\pm0.001}$ & -- & $\mathbf{0.945\pm0.000}$ & $67.9\pm0.5$ \\
\bottomrule
\end{tabular}
\end{table}

\paragraph{Space-filling-curve orderings.}
Table~\ref{tab:ordering} retrains MFNP on 2D Mat\'ern-$\nicefrac{3}{2}$
with four point orderings, one training seed each. Morton and lexicographic
orderings give $4\%$ and $9\%$ lower MSE than the Hilbert curve, lexicographic
ordering raises CRPS by $16\%$, and random ordering is $18\%$ worse.
We use the Hilbert curve because it keeps consecutive points adjacent
in the plane, as in prior Hilbert serialisation of irregular point
sets~\citep{liang2024pointmamba}, while the Morton curve jumps at each
quadrant boundary.

\begin{table}[!htbp]
\centering
\caption{Point orderings on 2D Gaussian Mat\'ern-$\nicefrac{3}{2}$ ($K{=}256$). Each ordering is retrained with the same budget and one training seed; the random row is the random-order run of Table~\ref{tab:ablation}. MSE and CRPS relative to the Hilbert run, mean over 10 evaluation seeds at $N_{\text{samples}}{=}128$.}
\label{tab:ordering}
\scriptsize
\setlength{\tabcolsep}{4pt}
\begin{tabular}{@{} l c c @{}}
\toprule
\textbf{Ordering} & \textbf{MSE / Hilbert}\,$\downarrow$ & \textbf{CRPS / Hilbert}\,$\downarrow$ \\
\midrule
  Hilbert       & $1.00$ & $1.00$ \\
  Morton        & $0.96$ & $\mathbf{0.98}$ \\
  Lexicographic & $\mathbf{0.91}$ & $1.16$ \\
  Random        & $1.18$ & $1.11$ \\
\bottomrule
\end{tabular}
\end{table}

\section{Solver and distillation}
\label{app:distillation}

On NOAA, the teacher generates the reflow couplings with $10$ Heun steps.
Table~\ref{tab:distillation} sweeps Euler and Heun integrators at $\{1, 15, 30\}$ steps on the reflow-distilled MambaFlowNP student across all four headline tasks. On 1D Mat\'ern-$\nicefrac{3}{2}$, $1$-step Euler matches the multi-step configurations within one SE on MSE and CRPS, and coverage moves by at most $0.005$ with more steps. On 2D Mat\'ern-$\nicefrac{3}{2}$, $1$-step Euler gives an MSE of $0.0167$, a CRPS of $0.0537$ and a coverage of $0.971$, against $0.0166$, $0.0537$ and $0.975$ at $15$ steps. On both NOAA tiers $1$-step Euler gives the lowest MSE, the lowest CRPS ($0.767$ against $0.808$ at $30$ steps on Utah, $0.681$ against $0.748$ at $15$ steps on California) and the coverage closest to nominal ($0.940$ and $0.945$). Heun-$1$ is within one SE of Euler-$1$ on MSE on 1D ($0.0112$ against $0.0111$) and on 2D ($0.0167$ for both). On the NOAA tiers it inflates MSE by $1.6$--$1.9{\times}$. The Euler 1-step row is therefore the deployment configuration used in every other table.

\begin{table}[!htbp]
\centering
\caption{Solver sweep on the reflow-distilled MFNP \emph{student} and the matched \emph{CFM-only teacher} (same architecture, hyperparameters, and evaluation; no reflow). Heun uses $2$ network function evaluations per step; time = per-episode ms, on one NVIDIA L40S. Mean $\pm$ SE over 10 evaluation seeds; bold = best per column within model block (within 1 SE). C2ST undefined on NOAA.}
\label{tab:distillation}
\scriptsize
\setlength{\tabcolsep}{4pt}
\begin{tabular}{@{} l ccccc @{}}
\toprule
 & \multicolumn{2}{c}{\textbf{Predictive Quality}} & \multicolumn{2}{c}{\textbf{Calibration Quality}} & \\
\cmidrule(lr){2-3} \cmidrule(lr){4-5}
\textbf{Config} & \textbf{MSE}\,$\downarrow$ & \textbf{CRPS}\,$\downarrow$ & \textbf{C2ST}\,$\rightarrow\!0.5$ & \textbf{Cov-95}\,$\rightarrow\!0.95$ & \textbf{Time\,(ms)}\,$\downarrow$ \\
\midrule

\multicolumn{6}{l}{\textit{1D Gaussian -- Mat\'ern-$\nicefrac{3}{2}$} ($K{=}128$)} \\[2pt]
\multicolumn{6}{@{}l}{\textbf{Teacher (pre-distillation)}} \\
\quad Euler 1-step  & $\mathbf{0.0112\pm0.0003}$ & $0.0506\pm0.0006$ & $0.621\pm0.005$ & $0.302\pm0.002$ & $\mathbf{28.5\pm0.1}$ \\
\quad Euler 15-step & $\mathbf{0.0111\pm0.0003}$ & $\mathbf{0.0397\pm0.0004}$ & $0.536\pm0.005$ & $0.903\pm0.001$ & $298.9\pm0.3$ \\
\quad Euler 30-step & $\mathbf{0.0111\pm0.0003}$ & $\mathbf{0.0396\pm0.0004}$ & $0.533\pm0.005$ & $0.923\pm0.001$ & $585.4\pm0.3$ \\
\quad Heun 1-step   & $\mathbf{0.0113\pm0.0003}$ & $0.0460\pm0.0005$ & $\mathbf{0.527\pm0.005}$ & $0.925\pm0.002$ & $48.4\pm0.3$ \\
\quad Heun 15-step  & $\mathbf{0.0111\pm0.0003}$ & $\mathbf{0.0396\pm0.0004}$ & $\mathbf{0.531\pm0.005}$ & $\mathbf{0.940\pm0.001}$ & $585.8\pm0.4$ \\
\quad Heun 30-step  & $\mathbf{0.0111\pm0.0003}$ & $\mathbf{0.0396\pm0.0004}$ & $\mathbf{0.531\pm0.005}$ & $\mathbf{0.940\pm0.001}$ & $(1.160\pm0.001){\times}10^{3}$ \\[2pt]
\multicolumn{6}{@{}l}{\textbf{Student (reflow-distilled)}} \\
\quad Euler 1-step  & $\mathbf{0.0111\pm0.0003}$ & $\mathbf{0.0398\pm0.0004}$ & $\mathbf{0.537\pm0.006}$ & $0.938\pm0.001$ & $\mathbf{29.5\pm0.2}$ \\
\quad Euler 15-step & $\mathbf{0.0111\pm0.0003}$ & $\mathbf{0.0396\pm0.0004}$ & $\mathbf{0.532\pm0.005}$ & $0.942\pm0.001$ & $299.1\pm0.7$ \\
\quad Euler 30-step & $\mathbf{0.0111\pm0.0003}$ & $\mathbf{0.0396\pm0.0004}$ & $\mathbf{0.532\pm0.005}$ & $0.942\pm0.001$ & $585.5\pm0.4$ \\
\quad Heun 1-step   & $\mathbf{0.0112\pm0.0003}$ & $0.0403\pm0.0004$ & $\mathbf{0.538\pm0.006}$ & $\mathbf{0.954\pm0.001}$ & $48.7\pm0.3$ \\
\quad Heun 15-step  & $\mathbf{0.0111\pm0.0003}$ & $\mathbf{0.0396\pm0.0004}$ & $\mathbf{0.532\pm0.005}$ & $0.943\pm0.001$ & $585.2\pm0.2$ \\
\quad Heun 30-step  & $\mathbf{0.0111\pm0.0003}$ & $\mathbf{0.0396\pm0.0004}$ & $\mathbf{0.532\pm0.005}$ & $0.943\pm0.001$ & $(1.157\pm0.002){\times}10^{3}$ \\
\midrule

\multicolumn{6}{l}{\textit{2D Gaussian -- Mat\'ern-$\nicefrac{3}{2}$} ($K{=}256$)} \\[2pt]
\multicolumn{6}{@{}l}{\textbf{Teacher (pre-distillation)}} \\
\quad Euler 1-step  & $\mathbf{0.0164\pm0.0005}$ & $0.0578\pm0.0006$ & $0.594\pm0.007$ & $0.669\pm0.001$ & $\mathbf{104.0\pm0.3}$ \\
\quad Euler 15-step & $\mathbf{0.0164\pm0.0005}$ & $\mathbf{0.0530\pm0.0005}$ & $\mathbf{0.554\pm0.005}$ & $\mathbf{0.962\pm0.001}$ & $(1.459\pm0.000){\times}10^{3}$ \\
\quad Euler 30-step & $\mathbf{0.0164\pm0.0005}$ & $\mathbf{0.0532\pm0.0005}$ & $\mathbf{0.553\pm0.005}$ & $0.970\pm0.001$ & $(2.909\pm0.001){\times}10^{3}$ \\
\quad Heun 1-step   & $\mathbf{0.0164\pm0.0005}$ & $0.0569\pm0.0006$ & $0.565\pm0.009$ & $0.915\pm0.001$ & $198.2\pm0.3$ \\
\quad Heun 15-step  & $\mathbf{0.0164\pm0.0005}$ & $\mathbf{0.0535\pm0.0005}$ & $\mathbf{0.552\pm0.005}$ & $0.976\pm0.000$ & $(2.913\pm0.000){\times}10^{3}$ \\
\quad Heun 30-step  & $\mathbf{0.0164\pm0.0005}$ & $\mathbf{0.0535\pm0.0005}$ & $\mathbf{0.552\pm0.005}$ & $0.976\pm0.000$ & $(5.819\pm0.001){\times}10^{3}$ \\[2pt]
\multicolumn{6}{@{}l}{\textbf{Student (reflow-distilled)}} \\
\quad Euler 1-step  & $\mathbf{0.0167\pm0.0005}$ & $\mathbf{0.0537\pm0.0005}$ & $\mathbf{0.554\pm0.005}$ & $\mathbf{0.971\pm0.000}$ & $\mathbf{103.6\pm0.3}$ \\
\quad Euler 15-step & $\mathbf{0.0166\pm0.0005}$ & $\mathbf{0.0537\pm0.0005}$ & $\mathbf{0.553\pm0.005}$ & $0.975\pm0.000$ & $(1.459\pm0.000){\times}10^{3}$ \\
\quad Euler 30-step & $\mathbf{0.0166\pm0.0005}$ & $\mathbf{0.0538\pm0.0005}$ & $\mathbf{0.553\pm0.005}$ & $0.975\pm0.000$ & $(2.908\pm0.000){\times}10^{3}$ \\
\quad Heun 1-step   & $\mathbf{0.0167\pm0.0005}$ & $0.0543\pm0.0005$ & $\mathbf{0.555\pm0.004}$ & $0.979\pm0.000$ & $198.6\pm0.3$ \\
\quad Heun 15-step  & $\mathbf{0.0166\pm0.0005}$ & $\mathbf{0.0538\pm0.0005}$ & $\mathbf{0.553\pm0.005}$ & $0.975\pm0.000$ & $(2.914\pm0.001){\times}10^{3}$ \\
\quad Heun 30-step  & $\mathbf{0.0166\pm0.0005}$ & $\mathbf{0.0538\pm0.0005}$ & $\mathbf{0.553\pm0.005}$ & $0.975\pm0.000$ & $(5.824\pm0.001){\times}10^{3}$ \\
\midrule

\multicolumn{6}{l}{\textit{NOAA Utah} ($K{=}222$)} \\[2pt]
\multicolumn{6}{@{}l}{\textbf{Teacher (pre-distillation)}} \\
\quad Euler 1-step  & $\mathbf{2.536\pm0.016}$ & $0.991\pm0.002$ & -- & $0.309\pm0.001$ & $\mathbf{28.26\pm0.08}$ \\
\quad Euler 15-step & $2.559\pm0.017$& $0.846\pm0.002$ & -- & $0.669\pm0.000$ & $331.77\pm0.15$ \\
\quad Euler 30-step & $2.571\pm0.017$ & $0.832\pm0.002$ & -- & $0.725\pm0.000$ & $656.49\pm0.19$ \\
\quad Heun 1-step   & $3.075\pm0.017$ & $1.451\pm0.001$ & -- & $\mathbf{0.996\pm0.000}$ & $49.71\pm0.06$ \\
\quad Heun 15-step  & $2.579\pm0.017$ & $\mathbf{0.813\pm0.002}$ & -- & $0.811\pm0.000$ & $658.35\pm0.17$ \\
\quad Heun 30-step  & $2.582\pm0.017$ & $0.817\pm0.002$ & -- & $0.793\pm0.000$ & $(1.310\pm0.000){\times}10^{3}$ \\[2pt]
\multicolumn{6}{@{}l}{\textbf{Student (reflow-distilled)}} \\
\quad Euler 1-step  & $\mathbf{2.464\pm0.016}$ & $\mathbf{0.767\pm0.002}$ & -- & $\mathbf{0.940\pm0.000}$ & $\mathbf{43.3\pm0.4}$ \\
\quad Euler 15-step & $\mathbf{2.471\pm0.015}$ & $0.809\pm0.002$ & -- & $0.771\pm0.000$ & $333.51\pm0.19$ \\
\quad Euler 30-step & $\mathbf{2.478\pm0.015}$ & $0.808\pm0.002$ & -- & $0.779\pm0.000$ & $656.77\pm0.10$ \\
\quad Heun 1-step   & $3.896\pm0.017$ & $1.192\pm0.001$ & -- & $0.988\pm0.000$ & $50.02\pm0.06$ \\
\quad Heun 15-step  & $2.483\pm0.015$ & $0.798\pm0.002$ & -- & $0.828\pm0.000$ & $659.34\pm0.16$ \\
\quad Heun 30-step  & $2.484\pm0.015$ & $0.802\pm0.002$& -- & $0.808\pm0.000$ & $(1.310\pm0.000){\times}10^{3}$ \\
\midrule

\multicolumn{6}{l}{\textit{NOAA California} ($K{=}486$)} \\[2pt]
\multicolumn{6}{@{}l}{\textbf{Teacher (pre-distillation)}} \\
\quad Euler 1-step  & $\mathbf{1.994\pm0.006}$ & $0.917\pm0.001$ & -- & $0.366\pm0.000$ & $\mathbf{97.95\pm0.18}$ \\
\quad Euler 15-step & $2.037\pm0.006$ & $0.803\pm0.001$ & -- & $0.671\pm0.000$ & $(1.373\pm0.000){\times}10^{3}$ \\
\quad Euler 30-step & $2.050\pm0.006$ & $0.792\pm0.001$ & -- & $0.715\pm0.001$ & $(2.740\pm0.000){\times}10^{3}$ \\
\quad Heun 1-step   & $2.238\pm0.005$ & $1.331\pm0.001$ & -- & $\mathbf{0.999\pm0.000}$ & $188.96\pm0.14$ \\
\quad Heun 15-step  & $2.055\pm0.006$ & $\mathbf{0.769\pm0.001}$ & -- & $0.800\pm0.000$ & $(2.738\pm0.000){\times}10^{3}$ \\
\quad Heun 30-step  & $2.060\pm0.006$ & $0.776\pm0.001$ & -- & $0.777\pm0.000$ & $(5.469\pm0.000){\times}10^{3}$ \\[2pt]
\multicolumn{6}{@{}l}{\textbf{Student (reflow-distilled)}} \\
\quad Euler 1-step  & $\mathbf{1.808\pm0.006}$ & $\mathbf{0.681\pm0.001}$ & -- & $\mathbf{0.945\pm0.000}$ & $\mathbf{67.9\pm0.5}$ \\
\quad Euler 15-step & $1.973\pm0.005$& $0.748\pm0.001$ & -- & $0.827\pm0.000$ & $(1.373\pm0.000){\times}10^{3}$ \\
\quad Euler 30-step & $1.978\pm0.005$ & $0.749\pm0.001$ & -- & $0.827\pm0.000$ & $(2.739\pm0.000){\times}10^{3}$ \\
\quad Heun 1-step   & $3.368\pm0.007$ & $1.026\pm0.001$ & -- & $0.850\pm0.000$ & $188.98\pm0.17$ \\
\quad Heun 15-step  & $1.986\pm0.005$ & $0.750\pm0.001$& -- & $0.832\pm0.000$ & $(2.738\pm0.000){\times}10^{3}$ \\
\quad Heun 30-step  & $1.985\pm0.005$ & $0.750\pm0.001$& -- & $0.830\pm0.000$ & $(5.469\pm0.000){\times}10^{3}$ \\
\bottomrule
\end{tabular}
\end{table}

\begin{figure}[!htbp]
\centering
\includegraphics[width=0.7\linewidth]{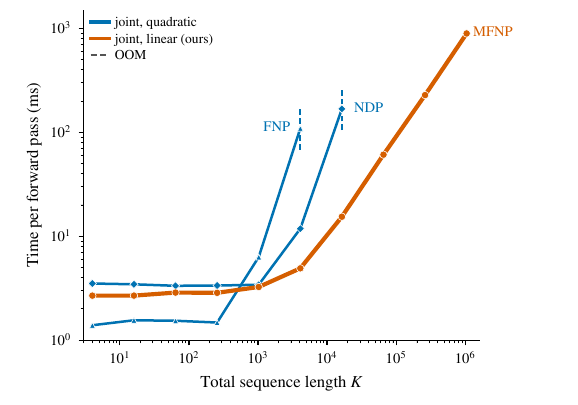}
\caption{Time per forward pass vs.\ total sequence length $K$ (log--log), 2D Gaussian Mat\'ern-$\nicefrac{3}{2}$, batch$=1$, \emph{1-NFE for all models} and matched ${\sim}600$k parameters with random initialisation. We plot per-call latency rather than the time per joint predictive draw because at 1-NFE only MFNP produces a usable sample; NDP and FNP need their full solver budget to be sampled from in practice. A short dashed line through a curve's last point means the next $K$ runs out of memory. With reflow's NFE budget equalised, NDP and FNP collapse onto MFNP's latency at $K\!\le\!1{,}024$ and diverge only once attention's $K^2$ cost dominates; FNP exhausts memory at $K{=}16{,}384$ and NDP at $K{=}65{,}536$.}
\label{fig:scaling-nfe1}
\end{figure}

\paragraph{Architecture vs.\ distillation at matched NFE.} The
headline scaling figure (Figure~\ref{fig:scaling}) leaves each baseline
at its native solver length, so MFNP's advantage there conflates
reflow distillation with the linear-$K$ encoder. To isolate the
two, we re-time NDP, FNP and MFNP at 1-NFE on the same 2D
Mat\'ern-$\nicefrac{3}{2}$ sweep with all three resized to
$\sim$600k parameters and random initialisation; forward-pass
latency depends only on architecture shape, so weights and trained
quality are immaterial here (Figure~\ref{fig:scaling-nfe1}). At
$K\!\le\!1{,}024$ all three sit on a $1.5$--$3.5$\,ms dispatch floor, so
the $34$--$90{\times}$ speedup observed at $K\le256$ (Table~\ref{tab:scaling}) is
attributable to reflow, not the encoder. The architectural
advantage emerges in the crossover regime (NOAA US, $K{=}5{,}580$),
where attention's $K^2$ cost begins to dominate, and is the binding
constraint beyond it. At 1-NFE FNP exhausts memory at $K{=}16{,}384$ and NDP at $K{=}65{,}536$, while MFNP scales to $K{=}10^{6}$. The two
contributions compose multiplicatively at NOAA US for the
$628$--$784{\times}$ end-to-end speedup reported in
Table~\ref{tab:noaa-main}.

\section{Boundary-adjacency diagnostic}
\label{app:boundary}

The Hilbert sort visits 2D points along a single space-filling curve, and at the curve's seams two points that are close in the plane can be placed far apart in the 1D order. Reversing a 1D sequence preserves all pairwise distances and index adjacencies, so $\pi_d^{\mathrm{rev}}$ does not relocate seams introduced by $\pi_d$; its role is to remove the causal-direction bias of the recurrence. The two distinct orderings $\pi_1$ and $\pi_2$ place seams in different locations, so a query whose neighbourhood is split under $\pi_1$ is typically contiguous under $\pi_2$. We test whether MFNP's predictive error concentrates on queries near these seams. Each query is ranked by its \emph{isolation}, defined as the larger of the Euclidean distances to its two immediate Hilbert-sort neighbours in the combined context--query set, minimised across the axis-transposed orderings $\pi_1$ and $\pi_2$. We then compare the mean squared error of the top decile (\emph{boundary}) against the bottom decile (\emph{interior}) over $256$ evaluation episodes per task. FNP, which uses full attention with no serialisation, serves as the no-sort control.

The diagnostic is run on 2D Gaussian Mat\'ern-$\nicefrac{3}{2}$ ($K{=}256$ uniform draws on $[0,1]^2$), NOAA Utah ($K{=}222$ stations), and NOAA California ($K{=}486$ stations). The synthetic task is the cleanest control, because a smooth isotropic stochastic process on a regular bounded domain isolates the diagnostic from the anisotropy, station clustering and discontinuous coastlines of the real-data tiers. Results are in Table~\ref{tab:boundary}.

On the synthetic task, the MFNP boundary/interior MSE ratio is $3.48$ and the FNP ratio is $3.00$. The two $95\%$ bootstrap confidence intervals overlap. On both NOAA tiers, MFNP and FNP yield comparable ratios ($1.40$ vs.\ $1.54$ on Utah; $1.43$ vs.\ $1.47$ on California), again with overlapping bootstrap intervals. The per-point Pearson and Spearman correlations between squared error and isolation remain below $0.18$ in every cell, with $r^{2}<0.04$; $11$ of the $12$ have $p<10^{-5}$ and one has $p{=}0.05$; this significance reflects the $n\geq 28{,}000$ sample size rather than effect size.

FNP, which never invokes the Hilbert sort, exhibits a boundary penalty of comparable magnitude to MFNP on every task. The penalty is therefore attributable to spatial extrapolation in regions with fewer informative context neighbours, not to the Hilbert serialisation. The synthetic task amplifies the absolute ratios while preserving the cross-architecture equivalence.

\begin{table}[!htbp]
\centering
\caption{Boundary-adjacency diagnostic. Boundary and interior MSE are the top and bottom deciles of query targets ranked by isolation score (distance to the two Hilbert-sort neighbours). NOAA values are on the quantile-normalised target scale, so are not comparable to the (\textdegree C)$^2$ MSE values in Table~\ref{tab:noaa-main}; the 2D Mat\'ern-$\nicefrac{3}{2}$ values are on the native model output scale. Ratios are dimensionless; $95\%$ CIs are from $2{,}000$ bootstrap resamples. Correlations are per-point squared error against isolation; $^{*}p<0.05$, $^{**}p<0.01$, $^{***}p<0.001$. \textsuperscript{\dag} our method.}
\label{tab:boundary}
\scriptsize
\setlength{\tabcolsep}{4pt}
\begin{tabular}{@{} l c c c c c @{}}
\toprule
\textbf{Model} & \textbf{Boundary MSE}\,$\downarrow$ & \textbf{Interior MSE}\,$\downarrow$ & \textbf{Ratio [95\% CI]} & \textbf{Pearson $r$} & \textbf{Spearman $\rho$} \\
\midrule

\multicolumn{6}{l}{\textit{2D Gaussian -- Mat\'ern-$\nicefrac{3}{2}$} ($K{=}256$)} \\[2pt]
  FNP                                     & $0.0189$ & $0.0063$ & $3.00$\,$[2.40,3.77]$ & $0.070^{***}$ & $0.123^{***}$ \\
  MFNP\textsuperscript{\dag}       & $0.0267$ & $0.0077$ & $3.48$\,$[2.74,4.30]$ & $0.064^{***}$ & $0.130^{***}$ \\
\addlinespace

\midrule

\multicolumn{6}{l}{\textit{NOAA Utah} ($K{=}222$)} \\[2pt]
  FNP           & $0.0308$ & $0.0200$ & $1.54$\,$[1.18,1.99]$ & $0.028^{***}$ & $0.075^{***}$ \\
  MFNP\textsuperscript{\dag} & $0.0260$ & $0.0185$ & $1.40$\,$[1.07,1.82]$ & $0.012$       & $0.059^{***}$ \\
\addlinespace

\midrule

\multicolumn{6}{l}{\textit{NOAA California} ($K{=}486$)} \\[2pt]
  FNP           & $0.0317$ & $0.0215$ & $1.47$\,$[1.36,1.60]$ & $0.034^{***}$ & $0.046^{***}$ \\
  MFNP\textsuperscript{\dag} & $0.0211$ & $0.0148$ & $1.43$\,$[1.31,1.55]$ & $0.019^{***}$ & $0.043^{***}$ \\
\bottomrule
\end{tabular}
\end{table}

\newpage
\section*{NeurIPS Paper Checklist}

\begin{enumerate}

\item {\bf Claims}
    \item[] Question: Do the main claims made in the abstract and introduction accurately reflect the paper's contributions and scope?
    \item[] Answer: \answerYes{}
    \item[] Justification: The four contributions enumerated in Section~\ref{sec:intro} (sub-quadratic joint-predictive NP, single-state context-compression diagnostic, lowest RMSE on every NOAA tier with $114$--$784{\times}$ speed-up, and $10^{6}$-point tractability) are each backed by a labelled table. Tables~\ref{tab:noaa-main} and~\ref{tab:synth-main} in the body report accuracy, and Tables~\ref{tab:ablation} and~\ref{tab:scaling} in the appendix report the compression and routing ablations and the scaling.

\item {\bf Limitations}
    \item[] Question: Does the paper discuss the limitations of the work performed by the authors?
    \item[] Answer: \answerYes{}
    \item[] Justification: Section~\ref{sec:conclusion} contains a Limitations paragraph; Section~\ref{sec:training} states the per-cell compute cap that excludes TNPA from the NOAA US tier; and the boundary diagnostic in Appendix~\ref{app:boundary} is included specifically to test a potential weakness of the Hilbert serialisation.

\item {\bf Theory assumptions and proofs}
    \item[] Question: For each theoretical result, does the paper provide the full set of assumptions and a complete (and correct) proof?
    \item[] Answer: \answerNA{}
    \item[] Justification: The paper contains no formal theoretical results. The single-state compression argument in Section~\ref{sec:theory} is empirical, with the ablation in Table~\ref{tab:ablation} (Appendix~\ref{app:ablations}) as supporting evidence.

    \item {\bf Experimental result reproducibility}
    \item[] Question: Does the paper fully disclose all the information needed to reproduce the main experimental results of the paper to the extent that it affects the main claims and/or conclusions of the paper (regardless of whether the code and data are provided or not)?
    \item[] Answer: \answerYes{}
    \item[] Justification: The architecture is specified in Section~\ref{sec:method} and Algorithms~\ref{alg:MFNP_detailed}, \ref{alg:MFNP_train_detailed}, and~\ref{alg:MFNP_inference_detailed} (Appendix~\ref{app:detailed-algs}); Section~\ref{sec:tasks} specifies task generation and NOAA splits; Section~\ref{sec:training} and~\ref{sec:protocol} specify the training and evaluation protocols; Appendix~\ref{app:hparams} (Tables~\ref{tab:hparams-1d}, \ref{tab:hparams-2d}, and \ref{tab:hparams-noaa}) lists the BO-selected per-cell architecture, learning rate, batch size, training-step counts, and inference NFE / guidance scale for every (model, task) cell.

\item {\bf Open access to data and code}
    \item[] Question: Does the paper provide open access to the data and code, with sufficient instructions to faithfully reproduce the main experimental results, as described in supplemental material?
    \item[] Answer: \answerYes{}
    \item[] Justification: Code, trained weights, and training and evaluation scripts for the NOAA experiments are released at \url{https://github.com/CosmoNaught/MambaFlowNP}, with instructions to reproduce the results. The NOAA GHCN-Daily data is public.

\item {\bf Experimental setting/details}
    \item[] Question: Does the paper specify all the training and test details (e.g., data splits, hyperparameters, how they were chosen, type of optimizer) necessary to understand the results?
    \item[] Answer: \answerYes{}
    \item[] Justification: Section~\ref{sec:training} specifies the optimiser family (AdamW), warmup-cosine schedule, EMA, Bayesian-optimisation sweep on validation MSE, and per-cell GPU-hour cap; Appendix~\ref{app:hparams} fills in the remaining details (weight decay $10^{-4}$, gradient clip $1.0$, $16{+}16$ BO trials per cell, and one sweep of $41$ to $50$ trials per task for MFNP on the synthetic tasks) and gives the per-cell architecture, learning rate, batch size, training steps, and inference settings in Tables~\ref{tab:hparams-1d}, \ref{tab:hparams-2d}, and~\ref{tab:hparams-noaa}; Section~\ref{sec:protocol} specifies the evaluation protocol ($10$ seeds, $N_{\text{samples}}{=}128$, denormalised scale, latency measurement).

\item {\bf Experiment statistical significance}
    \item[] Question: Does the paper report error bars suitably and correctly defined or other appropriate information about the statistical significance of the experiments?
    \item[] Answer: \answerYes{}
    \item[] Justification: The model-comparison, ablation and solver tables report mean $\pm$ one standard error across $10$ evaluation seeds (Section~\ref{sec:protocol}), with each seed covering $256$ synthetic episodes or $1{,}024$ NOAA episodes. The boundary diagnostic in Appendix~\ref{app:boundary} additionally reports $95\%$ bootstrap confidence intervals from $2{,}000$ resamples, and the remaining table captions state their own statistics.

\item {\bf Experiments compute resources}
    \item[] Question: For each experiment, does the paper provide sufficient information on the computer resources (type of compute workers, memory, time of execution) needed to reproduce the experiments?
    \item[] Answer: \answerYes{}
    \item[] Justification: All training and evaluation run on NVIDIA L40S GPUs and all timings are on a single L40S (Sections~\ref{sec:training} and~\ref{sec:protocol}), and each (model, task) cell is capped at $100$ GPU-hours covering training and all $10$ evaluation seeds (Section~\ref{sec:training}); per-episode inference latency is reported alongside every quality metric, and Table~\ref{tab:scaling} reports forward-pass latency vs.\ sequence length up to $K{=}10^{6}$.
    
\item {\bf Code of ethics}
    \item[] Question: Does the research conducted in the paper conform, in every respect, with the NeurIPS Code of Ethics \url{https://neurips.cc/public/EthicsGuidelines}?
    \item[] Answer: \answerYes{}
    \item[] Justification: The work uses publicly released NOAA temperature records and synthetic Gaussian-process simulations; no human subjects, scraped personal data, or sensitive attributes are involved, and the submission follows the NeurIPS Code of Ethics.

\item {\bf Broader impacts}
    \item[] Question: Does the paper discuss both potential positive societal impacts and negative societal impacts of the work performed?
    \item[] Answer: \answerNA{}
    \item[] Justification: This is foundational methodological work on amortised joint-predictive inference for sparse spatial fields. Positive applications (temperature, disease prevalence and rainfall mapping) are noted in Section~\ref{sec:intro}; we are not aware of a direct path to negative societal impact beyond those generic to scaling Bayesian inference.
    
\item {\bf Safeguards}
    \item[] Question: Does the paper describe safeguards that have been put in place for responsible release of data or models that have a high risk for misuse (e.g., pre-trained language models, image generators, or scraped datasets)?
    \item[] Answer: \answerNA{}
    \item[] Justification: The released artefact is a regression model for spatial interpolation trained on public weather-station temperatures; it does not generate images, language, or any output with identifiable misuse potential.

\item {\bf Licenses for existing assets}
    \item[] Question: Are the creators or original owners of assets (e.g., code, data, models), used in the paper, properly credited and are the license and terms of use explicitly mentioned and properly respected?
    \item[] Answer: \answerYes{}
    \item[] Justification: All baselines (FNP, TNPA, NDP, CNP/NP/ANP/CANP), Mamba-2, conditional flow matching, reflow distillation, and Hilbert-curve point-cloud routing are cited at point of use in Sections~\ref{sec:cfm}--\ref{sec:method}; the NOAA GHCN-Daily dataset is in the U.S.\ public domain and cited as~\citet{menne2012overview,menne2012ghcnd}.

\item {\bf New assets}
    \item[] Question: Are new assets introduced in the paper well documented and is the documentation provided alongside the assets?
    \item[] Answer: \answerYes{}
    \item[] Justification: The released code, weights and data are documented in the repository README.

\item {\bf Crowdsourcing and research with human subjects}
    \item[] Question: For crowdsourcing experiments and research with human subjects, does the paper include the full text of instructions given to participants and screenshots, if applicable, as well as details about compensation (if any)?
    \item[] Answer: \answerNA{}
    \item[] Justification: The paper involves no crowdsourcing or human-subjects research; experiments use synthetic Gaussian-process draws and public weather-station records.

\item {\bf Institutional review board (IRB) approvals or equivalent for research with human subjects}
    \item[] Question: Does the paper describe potential risks incurred by study participants, whether such risks were disclosed to the subjects, and whether Institutional Review Board (IRB) approvals (or an equivalent approval/review based on the requirements of your country or institution) were obtained?
    \item[] Answer: \answerNA{}
    \item[] Justification: No human-subjects research is performed.

\item {\bf Declaration of LLM usage}
    \item[] Question: Does the paper describe the usage of LLMs if it is an important, original, or non-standard component of the core methods in this research? Note that if the LLM is used only for writing, editing, or formatting purposes and does \emph{not} impact the core methodology, scientific rigor, or originality of the research, declaration is not required.
    \item[] Answer: \answerNA{}
    \item[] Justification: LLMs were used only for prose editing and code-organisation assistance; they are not part of the core method (a Mamba-2 state-space scan trained with conditional flow matching) and do not affect the scientific rigour or originality of the work, so per the question's own guidance no declaration is required.

\end{enumerate}

\end{document}